\documentclass[journal]{IEEEtai}

\usepackage[colorlinks,urlcolor=blue,linkcolor=blue,citecolor=blue]{hyperref}

\usepackage{color,array}

\usepackage{graphicx}

\usepackage{algorithm}%
\usepackage{algorithmicx}%
\usepackage{algpseudocode}%
\usepackage{adjustbox}
\usepackage{amsmath}
\usepackage{multirow}
\usepackage{subcaption}
\usepackage{amssymb}

\begin{document}

\title{Learning 3D MRI Representations with Generative Adversarial Networks} 

% \author{Anonymous Authors}
\author{Sujeet Lahane, Sapna Mishra, Akash Suryavanshi, Tapan Gandhi, and Bharat Biswal
% Submitted for review 09 July 2026. 
\thanks{This work was supported by the Ministry of Electronics and Information Technology (MeitY), Government of India, under grant 4(16)/2019-ITEA and the Cadence Chair Professor fund awarded to Prof. Tapan Gandhi.}
\thanks{Sujeet Lahane was with Department of Mechanical Engineering, IIT Delhi, New Delhi, 110016 India. He is now with Department of Computer Science, IIT Delhi, New Delhi, 110016 India (e-mail: sujeetlahaneqrs@gmail.com).}
\thanks{Sapna Mishra was with Department of Electrical Engineering, IIT Delhi, New Delhi, 110016 India. She is now with Department of Biomedical Engineering, NJIT, Newark, NJ 07102 USA (e-mail: mishra.sapna94@gmail.com).}
\thanks{Akash Suryavanshi is with Department of Computer Science, IIT Delhi, New Delhi, 110016 India (e-mail: akash.iitd@gmail.com).}
\thanks{Tapan Gandhi is with Department of Electrical Engineering, IIT Delhi, New Delhi, 110016 India (e-mail: tgandhi@iitd.ac.in).}
\thanks{Bharat Biswal is with Department of Biomedical Engineering, NJIT, Newark, NJ 07102 USA (e-mail: biswal@njit.edu).}}
% \thanks{This paragraph will include the Associate Editor who handled your paper.}}

\maketitle

\begin{abstract}
Learning representations of visual images is a challenging task. It requires extracting abstract patterns from high-dimensional data without explicit supervision. This becomes pronounced with 3D MR imaging due to sparse features, higher dimensional inputs, and smaller datasets. Many state-of-the-art representation learning algorithms successfully rely on transformers and diffusion-based architectures. However, these methods struggle with smaller datasets. In effect, solution methods relax the problem by slicing 3D volumes and solving the problem for 2D medical imaging, resulting in a loss of spatial 3D information. We present R2RG, a representation learning method based on Convolutional Neural Networks (CNNs), with Generative Adversarial Networks (GANs) honing the reconstruction task. Our key insights lie in exploiting architectural biases in 3D CNNs, combined with data-efficient auto-encoding GANs and a composite loss function to enhance reconstruction fidelity and stabilize training. Our experiments on three standard datasets and a locally collected dataset show that R2RG not only achieves superior performance but also generalizes to heterogeneous and smaller datasets. Furthermore, representations learned with R2RG outperform those learned with contrastive-based representation learning methods on Parkinson's and Alzheimer's disease classification. Our work demonstrates that reconstruction-based representation learning is a promising direction for 3D medical imaging under limited data availability.
\end{abstract}

\begin{IEEEImpStatement}
Representation learning is central to modern decision-making systems and has seen significant efforts in natural image and language analysis. Yet, it remains largely underexplored for 3D medical imaging, primarily due to scarce data and sparse features, despite its potential to benefit early disease detection and diagnosis. We combine and exploit data-efficient methods in deep learning and present R2RG as a framework to address these challenges. Our work shows that reconstruction-based representation learning has potential applications for automating clinical diagnoses with minimal labeled data. We hope our work encourages further research towards 3D medical image representation learning.
\end{IEEEImpStatement}


\section{Introduction}

Recent advances in deep neural networks have led to impressive achievements across a range of computer vision tasks, including image classification, object detection, and segmentation. Transformers and Diffusion based architectures have gained significant popularity due to their ability to scale with large datasets and achieve superhuman performance. 
However, due to lack of data, these methods do not scale well in medical imaging, with the best of these methods performing subpar compared to Convolutional Neural Network (CNN) based methods \cite{vit-ae}. These methods hence try to decompose the problem by slicing 3D volumes into 2D images \cite{2ddiffusions,2dtransformers}. This does not provide the method with spatial relations within a 3D volume.
 
Magnetic Resonance Imaging (MRI) is a pivotal modality in clinical practice due to its superior soft-tissue contrast. Yet, the sheer volume and complexity of MRI data inflict a significant cognitive load on radiologists, leading to delays and inter-observer variability that can impact patient care. To alleviate this burden, a proliferation of CNN-based tools has emerged, ranging from automated tumor detection to early-stage disease prediction, demonstrating impressive accuracy and reproducibility in controlled research settings \cite{fconn,ihc,dlseg,segment,hrest,dldisease,dltumor}. However, a key challenge remains: the limited availability of labeled 3D medical data, which poses a significant obstacle to fully leveraging CNNs in this domain \cite{data_scars}. To overcome this bottleneck, a proposed solution in the general computer vision community is in the form of representation learning.

Image representation learning corresponds to encoding images into a lower-dimensional space while preserving meaningful abstract information. Effective image representations can significantly improve the performance of machine learning models by capturing essential features and patterns within the data. This area is largely influenced by transformer-based methods, which are not very useful for our case. On the other hand, Generative Adversarial Networks (GANs) can generate synthetic data to enrich training sets with fewer examples\cite{alGAN}. Though, they are not inherently designed to produce or learn representations of the input datum.

\begin{figure}[t]
    \centering
    \includegraphics[width=79mm]{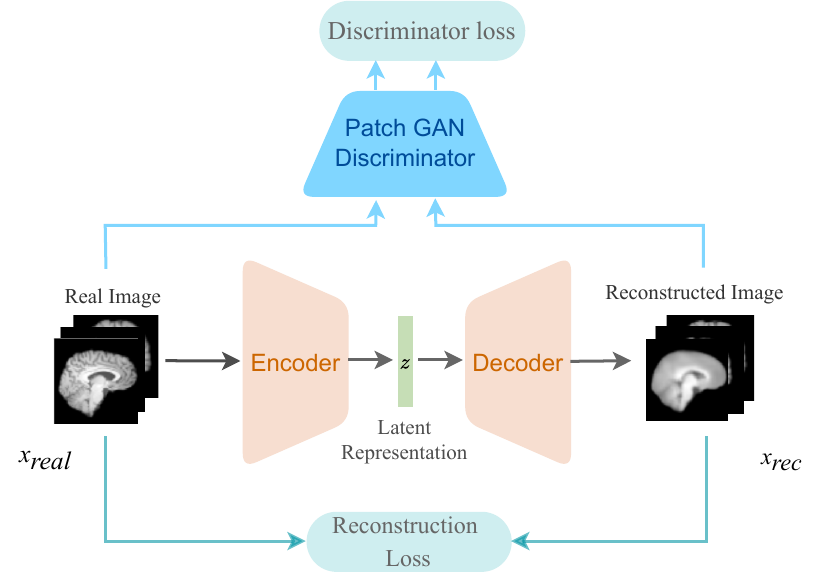}
    \caption{The proposed GAN-based reconstruction method, where a 3D volume $x_{real}$ is passed through the encoder block to get the representation vector \textit{z}. The vector $z$ is passed through the decoder to get the reconstructed output $x_{rec}$, which is used to evaluate reconstruction losses and distributional shift.}
    \label{fig:Fig_1}
\end{figure}

These challenges led to the exploration of learning image representations through reconstruction. Reconstruction-based representation learning seeks to encode images into a compact, abstract latent space by training networks to reproduce their own inputs from latents, ensuring no direct supervision and making it well-suited when labeled data is scarce. For instance, Pinapatruni \emph{et al.} \cite{NeRP} leverages reconstruction techniques to learn representations that generalize across various imaging modalities, demonstrating success even with limited data. Exploiting the reconstruction capabilities of deep neural networks has proven beneficial for addressing data scarcity, particularly in the context of learning 2D medical image representations \cite{recon_dl}. When dealing with MRI volumes, the objective extends to representing the entire 3D volume through a latent vector while retaining clinically relevant features. 

Given the data scarcity and complexity of methods such as contrastive learning \cite{contr}, we focus on learning representations through reconstruction using autoencoders. In this work, we propose R2RG, a novel self-supervised representation learning method, which aims to learn 3D MRI representation through reconstruction (see Fig. \ref{fig:Fig_1}).
Our method encodes volumetric data into latent representations to preserve spatial correlations in three-dimensional space. R2RG integrates GAN-based volume distribution modeling with an autoencoder architecture that enhances the quality of the representation in the latent space and mitigates the issue of data scarcity. We employ a PatchGAN-style discriminator to provide spatially localized feedback, helping expedite learning reconstruction. Furthermore, we introduce a composite loss function that combines $L_1$ and $L_2$ losses for reconstruction fidelity, a Lipschitz regularizer for training stability, and a distributional shift penalty to better model the spatial feature distribution \cite{gan}. 

We show that dense feedback to the generator through a patch-based discriminator results in a significant advantage in reconstruction. 
While many previous works use either the $L_1$ or $L_2$ norm as the sole objective, we find that a balanced combination leverages their complementary advantages, leading to overall improvement in reconstruction and representations. Finally, we show that improved reconstruction leads to better performance on the downstream task.

\section{Background and Problem Statement}
\label{sec2}

\subsection{Generative Adversarial Networks}
GANs introduce adversarial learning where a generator and a discriminator engage in a zero-sum game. A discriminator discriminates between real and generator augmented (fake) images, while generator tries to minimize distinguishability between real images and its generation \cite{gan}. Due to their capability of generating high fidelity outputs, GANs have gathered attention in tasks such as data augmentation, image-to-image translation, and art generation \cite{ganaugmed,cyclegan,ganart}. Additionally, GANs were also explored for representation learning, where image features are learned in the penultimate or former layers of the Generator or Discriminator \cite{rgan,rgan2}. GANs have found successful applications in medical imaging for data augmentation, cross-modal synthesis, and super-resolution \cite{ganmed,2D3D,2Dcrm,vtgan}. Their capacity to model complex distributions and produce anatomically plausible outputs makes them especially appealing for learning robust representations in clinical imaging applications.

\subsection{Representation Learning}
Representation learning mitigates the reliance on extensive \textit{annotated} datasets by first pre-training models to capture salient features from abundant unlabeled data. Through unsupervised objectives such as reconstruction, contrastive losses, or predictive tasks. A network learns to encode raw inputs into a compact latent space that preserves essential structure and semantics \cite{repr}. Once this rich embedding is established, only a small labeled subset is necessary to fine-tune a downstream task specific classifier or decoder, since the pre-trained representations already encapsulate meaningful information required for downstream tasks \cite{bert}. By decoupling feature extraction from label-intensive training, representation learning thus enables models to generalize effectively with fewer annotated examples, enhancing both efficiency and robustness in domains where expert labeling is costly and limited \cite{feat}.

SimCLR and BYOL are self-supervised learning methods designed to learn such useful data representations without the need for labeled data \cite{simclr,byol}. SimCLR uses contrastive learning with positive and negative sample examples to steer latent space, while BYOL relies on self-distillation, only relying on positive examples, and uses a ``target" network \cite{targetnetsdqn} and exponential moving averaging \cite{targetpolyak} in parameters to stabilize the training. These methods have demonstrated effective representation learning for natural images, and similar ideas have been extended with transformer architecture \cite{dino}. However, these methods have also shown performance degradation with reduction in dataset sizes \cite{contr-lim}, which is a primary concern in 3D medical imaging.

An alternative form of representation learning, reconstruction-based representation learning, emerged with autoencoders, which aimed to learn compact representations by minimizing the reconstruction error \cite{cdim}. This approach has since evolved and has proved particularly beneficial for problems with limited dataset sizes. In medical imaging, although reconstruction has been extensively explored, especially in tasks such as super-resolution, these methods often rely on skip connections that bypass the latent space, diluting its latent representational richness \cite{vaevq,2D3D}. As a result, the latent vector may not fully capture the volumetric or semantic information of the input. A commonly adopted workaround involves decomposing 3D volumes into 2D slices, thereby increasing data availability and simplifying training procedures \cite{2Dtrf, 2Dcrm}. This strategy has gained popularity in various 2D reconstruction frameworks \cite{2DTr, 2DSR,2Dtr3D, 2ddiffusions}; however, it inherently disrupts spatial continuity and loses inter-slice contextual information critical for volumetric understanding. In contrast, approaches to direct 3D reconstruction remain comparatively limited and rely predominantly on 3D CNNs \cite{recon_dl}. Modern transformers and diffusion-based methods show superhuman results in many tasks when presented with internet-scaled data. But perform poorer than CNNs in scarce data problems and are expensive in training \cite{diff-data,vit-cnn}. Moreover, with diffusion, it is often unclear which part of their architecture produces latent representation vectors. Due to such problems, transformers and diffusion-based works are largely limited to 2D medical imaging \cite{2DTr,2Dtr3D,2diff,2difrec}.

Methods such as $\alpha$-WGAN \cite{alGAN} focus on data augmentation; however, they often underperform in terms of reconstruction fidelity and the training is computationally expensive. In addition, $\alpha$-WGAN learns a latent space distribution for data augmentation rather than individual representations for MRI volumes. Contrastingly, 3D-VAEs have demonstrated potential due to their ability to learn continuous latent spaces and their architecture focusing specifically on reconstructions, but their performance typically lags behind their 2D counterparts, often producing blurry or oversmoothed outputs \cite{vae3d}.

\subsection{Problem Statement}

Given 3D MRI samples, we aim to learn an encoder $E$ which condenses a high-dimensional sample $x$ into a representation $z$ in a lower-dimensional latent space. Representation, $z$, shall capture relevant information in $x$ without any supervision. Thus, downstream tasks can operate directly on the latent representation, $z$, thereby reducing the learning overhead and lowering the amount of labeled data and training time required.

Formally, given a dataset, 
\[\mathcal{D}=\{x_i | x_i\in \mathbb{R}^{H\times W\times D}\}_{i=1}^m\] our goal is to learn a mapping,
\[E:\mathbb{R}^{H\times W\times D}\to \mathbb{R}^d, \text{with } d\ll {H\times W\times D}\] \[\text{such that, }\max_{E} \mathbb{E}_{x\sim\mathcal{D}}\mathcal{S}(x, E(x))\] where, $\mathcal{S}$ represents a notion of similarity between input arguments. Defining $\mathcal{S}$ objectively is challenging, the most common approximations include metrics derived from contrastive loss or reconstruction efficacy.

\section{Methodology}
\label{sec3}

Building on the success of large-scale machine learning applications in domains such as natural language processing and computer vision, it has become evident that learning effective data representations is also a crucial step forward in medical imaging \cite{repr, bert}. However, a significant barrier to progress in this direction is the data-intensive nature of modern deep learning models, which contrasts sharply with the limited availability of labeled data in medical imaging \cite{medrepr, medrepe}. In this section, we discuss the proposed method in detail. R2RG leverages a reconstruction-based learning strategy to encode 3D MRI volumes. By focusing on learning through reconstruction, R2RG minimizes the dependence on labeled datasets while effectively capturing meaningful volumetric representations.

\subsection{Network Architecture}
In the autoencoder paradigm, the network is trained to reconstruct its input by encoding an image into a lower-dimensional latent representation and subsequently decoding its latent representation back to the image. Through this process, the network learns to encode data while minimizing reconstruction error. On the other hand, a GAN, known for its ability to generate high-fidelity images, typically with a random noise passed to a decoder, plays a zero-sum game with the discriminator. To leverage the benefits of both methods, we fuse them together (illustrated in Fig.~\ref{fig:Fig_1}), hence R2RG is designed to learn representations that could reconstruct high-quality MRI volumes. The proposed architecture is an autoencoding GAN, where the generator of the GAN is a 3D encoder-decoder based reconstruction module, along with a novel 3D PatchGAN-based discriminator. PatchGAN class of discriminators improves on vanilla discriminators. Vanilla discriminators use a scalar element as feedback, hence do not provide the generator with sufficient information to improve on \cite{pgan}. On the other hand, PatchGAN provides a patch based detailed feedback. Thus, such low-level patch-based feedback helps speed up training and rapidly improves generator performance.

\begin{figure*}[t]
    \centering
    \begin{subfigure}{.92\textwidth}
        \centering
        \includegraphics[trim=12mm 3mm 8mm 6mm,clip,width=\textwidth]{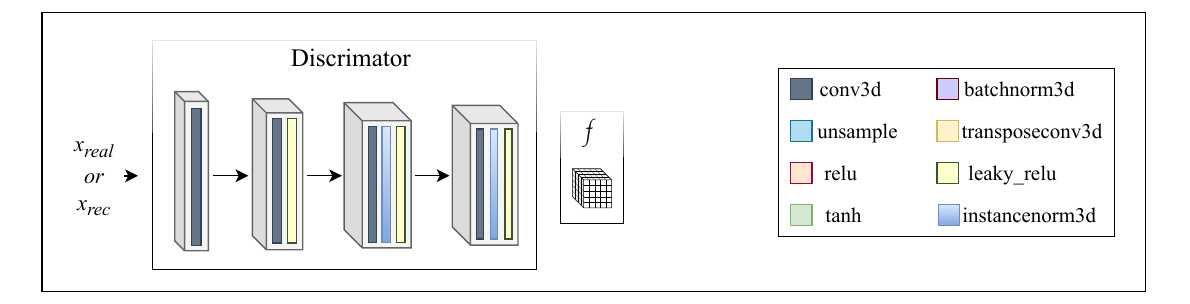}
        \caption{}
        \label{fig:Fig_2a}
    \end{subfigure}
    \begin{subfigure}{.93\textwidth}
        \centering
        \includegraphics[trim=8mm 10mm 8mm 3mm,clip,width=\textwidth]{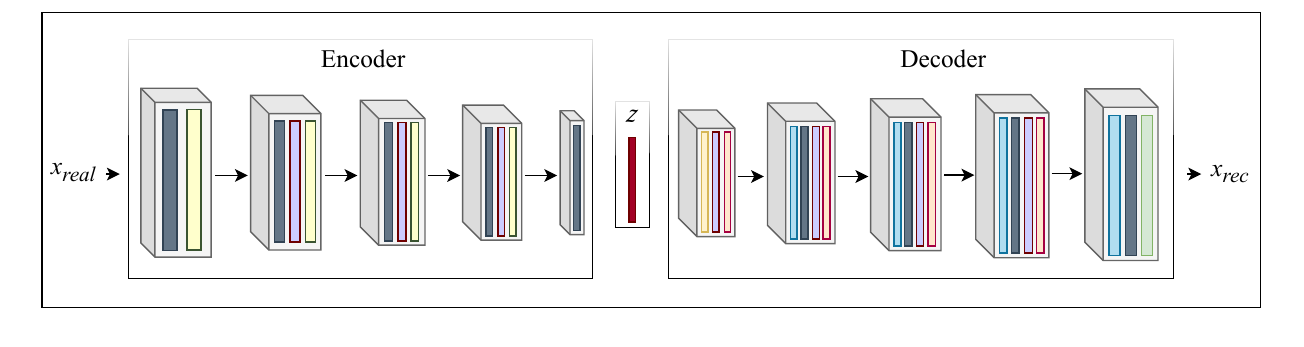}
        \caption{}
        \label{fig:Fig_2b}
    \end{subfigure}
    \caption{Modular illustration of the proposed R2RG network, where (a) shows the working of the PatchGAN Discriminator, which outputs a three-dimensional tensor \textit{f} as feedback to the input volume $x_{real}$ or $x_{rec}$, (b) shows the details of the Generator in our GAN network, which encompasses an encoder and a decoder. The encoder compresses the input volume ${x_{real}}$ into the representation vector \textit{z}, the decoder takes $z$ as input and reconstructs $x_{rec}$ such that $L_{RL}$ is minimized.}
    \label{fig:Fig_2}
\end{figure*}

The illustration of the proposed architecture is shown in Fig. \ref{fig:Fig_2}. Our 3D encoder network encompasses five sequential convolutional blocks, each with a kernel size of 4$\times$4$\times$4. This encodes the input volume into a latent space vector, $z$, of size 1000. The decoder network consists of a transpose convolution layer with kernel size 4$\times$4$\times$4 followed by four upsample+convolution blocks with kernel size 3$\times$3$\times$3. The upsample+convolution setup is preferred as it is less susceptible to the checkerboard artifacts as discussed in \cite{checkerboard} and uses fewer parameters. On the final layer, the $tanh$ activation function keeps the values of reconstructed volume within range of (-1, 1), which are then standardized to (0, 1) similar to input volume, to further feed into the discriminator. The discriminator network inspired by PatchGAN \cite{pgan} outputs a tensor, $f_{fake} = D(x_{\text{rec}})$ or $f_{real} = D(x_{\text{real}})$ of size 8$\times$8$\times$8 for reconstructed volume, and real volume as inputs respectively. The discriminator tries to maximize the distributional shift between outputs of real volumes and reconstructed volumes. Discriminator architecture comprises a feature-map block with a convolution layer and kernel of size 7$\times$7$\times$7, followed by three contracting blocks, each with a convolution layer and kernel of 4$\times$4$\times$4. Each convolutional block in the decoder is succeeded by instance normalization and leaky-ReLU activation for stable training, as seen in Fig. \ref{fig:Fig_2}.

The generator block consists of an encoder and a decoder. The encoder takes real volume as input and generates the representation vector $z$, which is then decoded by the decoder to reconstruct the volume. It should be noted that we do not connect the encoder and decoder by skip connections with the idea of keeping $z$ the only source of information passed to the decoder. The output of the generator, a reconstructed volume, is evaluated against the input volume for distributional shift and reconstruction loss. These volumes are further passed to the discriminator, which provides feedback, $f_{fake}$, for the zero-sum game of GAN.

From representation learning perspective comparing to existing methods, SimCLR learns to maximize agreement in the space immediately transformed from representation space \cite{simclr}, BYOL learns to maximize agreement in the prediction space \cite{byol}, our method can be thought of as learning representations to maximize agreement in the volume space.

\subsection{Loss Function}

In the proposed network, the reconstructed volume $x_{rec}$, the output of the decoder, is treated as a fake volume, and the corresponding input of the encoder, $x_{real}$, is considered as a real volume by the discriminator. To train the discriminator, we use the Wasserstein loss \cite{wgan} with an added gradient penalty term. The Wasserstein loss enables the discriminator to maximize the distribution distance between its output on real instances and on fake instances, given as:
\begin{equation}
    \mathbb{E}_{x_{rec}}[D(x_{\text{rec}})] - \mathbb{E}_{x_{\text{real}}}[D(x_{\text{real}})]
\end{equation}
 
The incorporated gradient penalty term, ${L_{GP}}$, enforces 1-lipschitz continuity on gradients of the discriminator, hence stabilizes training \cite{wgangp}. The gradient penalty is calculated as 
\begin{equation}
{L_{GP}} =
\mathbb{E}_{x_{rec}}[(\lVert \nabla_{x_{rec}}D(x_{rec}) \rVert_{\ell_2}-1)^2].
\end{equation}

Together, the Wasserstein loss evaluated on $x_{\text{real}}$ and $x_{\text{rec}}$ and the gradient penalty term $L_{GP}$ form the discriminator loss function $L_D$, defined as:

\begin{equation}
    L_D = \mathbb{E}_{x_{\text{rec}}}[D(x_{\text{rec}})] - \mathbb{E}_{x_{\text{real}}}[D(x_{\text{real}})] + \lambda_{GP} L_{GP}
\end{equation}
where $\lambda_{GP}$ regulates the weight of the gradient penalty in the discriminator loss.

Autoencoder-based generator networks often produce blurry reconstructed images, while standard GAN generators can suffer from mode collapse and unstable training. The $L_1$ loss, though it stabilizes training, tends to produce blurry images %converge to a local optimum, 
whereas the $L_2$ loss, which encourages output fidelity, destabilizes training. To mitigate these issues, our proposed method employs a combination of $L_1$ and $L_2$ losses for training the generator, achieving a balance between stability and fidelity. 

Furthermore, we incorporate the output of the discriminator in the reconstructed images, $D(x_{\text{rec}})$, into the loss function of the generator. The generator is incentivized to maximize $D(x_{\text{rec}})$, which in turn tries to maximize the discriminator's output for its fake instances, thereby increasing the discriminator's loss. The discriminator training, hence, acts as feedback for the generator training, leading to high-quality reconstructions. The loss function used to train the generator, here termed as reconstruction loss, is calculated as the combination of $L_1$, $L_2$, and discriminator feedback. It is represented as $L_{RL}$, and the formulation is given below.
\begin{equation}
    L_{RL} = -\mathbb{E}_{x_{\text{rec}}}[D(x_{\text{rec}})] + \lambda_1 \lVert x_{\text{real}} - x_{\text{rec}} \rVert_{\ell_1} + \lambda_2 \lVert x_{\text{real}} - x_{\text{rec}} \rVert_{\ell_2}
\end{equation}
Here, $\lambda_1$ and $\lambda_2$ are parameters that control the contributions of the $L_1$ and $L_2$ losses, respectively.

\subsection{Classification}
\label{321}
We further evaluate the representation capabilities of our model on a downstream classification task with a shallow neural network classifier, similar to \cite{simclr,byol}. For MRI classification tasks, deep learning methods are best evaluated on diseases and datasets where the condition exhibits detectable effects on MRI scans. Previous research indicates that Parkinson’s and Alzheimer's disease are associated with visible markers on brain MRI \cite{ppmiPD}. Hence, we make use of PPMI, ADNI1 datasets for the classification task to validate the proposed representation learning method. By leveraging these subtle anatomical variations, the classifier shall be able to assess the discriminative quality of the learned representations. The classification module we use here is a single hidden layer shallow neural network (16 - 32 hidden units) classifier with a dropout layer for generalization. We first pass the MRI volumes to our encoder trained on the corresponding train sets, and then pass the output representations as input to the classifier. Here, the encoder is treated as a constant transformation function, mapping input volumes to a representation vector.

\section{Datasets and Evaluation Metrics}
\label{sec4}
\subsection{Training Data} 
For the reconstruction task, we employ experiments with three publicly available datasets with healthy controls. The Amsterdam Open MRI Collection (AOMIC) \cite{pomic}, Alzheimer's Disease Neuroimaging Initiative (ADNI1) \cite{adni}, and Parkinson's Progression Markers Initiative (PPMI)\footnote{Data used in the preparation of this article were obtained on 2024-05-30 from the PPMI database (\url{https://www.ppmi-info.org/access-data-specimens/download-data}), RRID: SCR\_006431. For up-to-date information on the study, visit \url{http://www.ppmi-info.org}.} \cite{ppmi} datasets. AOMIC contains T1-weighted brain MRI data from 928 healthy patients, each contributing to various paradigms, hence a total of 2784 volumes \cite{aomic}. In our setup, 2131 volumes are included in the training set, and the remaining volumes are added to the test set.

We further validate our model on the PPMI \cite{ppmi} and ADNI datasets. The PPMI is a collaborative dataset and samples arrive from various sources, hence, it lacks consistency across the examples. We take a subset of the dataset with 289 Parkinson's Disease (PD) and 277 healthy controls (HC) to maintain the ratio of PD and control examples. We divide the dataset into train-validation-test sets as 191, 43, 43 for PD and 199, 46, 44 for HC samples. ADNI \cite{adni} dataset consists of structural T1-weighted Brain MRI data from 957 patients. This dataset is classified into 421 patients with Mild Cognitive Impairment (MCI), 198 with Alzheimer's Disease (AD) and 238 were Cognitively Normal (CN). For our experiments, the dataset is divided such that training contains 294 MCI, 138 AD and 236 CN samples. The remaining samples are equally divided into validation and test sets.

\subsubsection{COVID Data} 
To further confirm the generalization of our model, we tested our model with a locally collected dataset of 51 healthy controls and 77 COVID-19 recovered patients aged 18 to 65 years. The COVID-19 patients were scanned within six months of their recovery. Patients with pre-existing neurological conditions or who required ventilation during their COVID-19 infection were excluded.

A 3T GE scanner with a 32-channel head coil was used to capture T1-weighted (T1w) Brain MRI images in 3D imaging mode with the fast BRAVO sequence. The image parameters were: inversion time (TI) of 450 ms, a flip angle of 12 degrees, a field of view (FOV) spanning 256 mm $\times$ 256 mm, 152 slices with a slice thickness of 1.00 mm, and an isotropic resolution of 1 mm.

For evaluations, COVID data is split in 7:3 ratio for train-test sets. The dataset is largely included in this study to demonstrate the capability of our method on an off-the-shelf, local dataset.

\subsection{Training Process}
\label{sec42}

Data samples of all four datasets were preprocessed to strip the non-brain regions with the help of the Brain Extraction Tool from the \textit{fsl} software \footnote{\url{https://fsl.fmrib.ox.ac.uk/fsl/fslwiki/FSL}}. Further, the volumes were padded, average-pooled and resized to a tensor of shape 64$\times$64$\times$64 for computational efficiency and uniformity in data. 

The generator network, fed with pre-processed volumes, performs the reconstruction task. The encoder and decoder network, treated as a generator, takes two training steps against a training step of a discriminator due to its comparatively flatter learning curve. The discriminator is backpropagated with $L_D$ and is kept frozen while backpropagating $L_{RL}$ through the generator (see Algorithm \ref{alg:train}). We first train our network from scratch on the AOMIC dataset, and evaluate on the AOMIC dataset. Then we use this pretraining to fine-tune on the smaller datasets for their evaluation.

We implemented all the methods using PyTorch 2.1 and trained with an NVIDIA A30 GPU. The proposed method was trained with Adam optimizer, learning rate of 0.0001 and a batch size of 8. Training SimCLR and BYOL requires augmenting data, which is achieved with a Python library, TorchIO's 3D augmentation techniques. For classification, learning rate was fixed at 0.0003 with a batch size of 32.

\begin{algorithm}
\caption{Pseudo-code for training R2RG}
\label{alg:train}
\begin{algorithmic}
    \State \textbf{Input:} Preprocessed data $X_{train}$, number of training iterations $iters$
    \State Initialize Discriminator $D$, Generator $R$ comprised of Encoder $E$ and Decoder $G$ ($R:=G$$\circ$$E$)
    \For{$i \gets 1$ to $iters$}
        \State sample mini-batch $x \sim X_{train}$
        \State $z \gets E(x)$
        \State $x_{r} \gets G(z)$
        % \State $f \gets D(x_{rec})$
        \State $L_{C} = \lambda_1 \|x-x_r\|_{\ell_1} + \lambda_2 
        \|x-x_r\|_{\ell_2}$
        \State $x_{r}$.detach()
        \State $L_{RL} \gets - \mathbb{E}_{x_r}[D(x_{r})] + L_C$
        \State $R.update(L_{RL})$ \Comment{Applied twice}
        % \State $R.update(loss_R)$
        \State $L_D \gets \mathbb{E}_{x_r}[D(x_{r})] - \mathbb{E}_{x}[D(x)] + \lambda_{GP}(\lVert \nabla D \rVert_{\ell2}-1)^2$
        \State $D.update(L_D)$
    \EndFor
    \State \Return{$E$}
\end{algorithmic}
\end{algorithm}

\subsection{Evaluation Metrics} 
We evaluate the reconstruction fidelity at the test set of all the datasets using the Mean Structural Similarity Index Measure (MSSIM) \cite{ssim} and the Peak Signal to Noise Ratio (PSNR), the standard metrics for reconstruction evaluation. 
MSSIM of a reconstructed volume (Y) corresponding to input real volume (X) with number of windows $M$, is evaluated as,
\begin{equation}
    \text{MSSIM}(X,Y) = \frac{1}{M} \sum_{j}^{M} \frac{(2\mu_{x_{j}}\mu_{y_{j}} + C_1) (2\sigma_{x_{j}y_{j}} + C_2)}{(\mu_{x_{j}}^2 + \mu_{y_{j}}^2 + C_1) (\sigma_{x_{j}}^2 + \sigma_{y_{j}}^2 + C_2)}
\end{equation}
where, $\mu_{x_{j}}$ and $\sigma_{x_{j}}$ denotes mean and variance across a window, and $\sigma_{xy_j}$ is a covariance across two different windows. Standard values are used for the constants $C_1$ and $C_2$ \cite{ssim}.

And under standard formulation we evaluate PSNR as, 
\begin{equation}
    \text{PSNR}(X,Y) = A \log\left({max(X)}\right) - B \log{\left(\frac{1}{|X|}\sum (X-Y)^2\right)} %= B \log{\left(\frac{1}{|X|}\sum (X-Y)^2\right)}
\end{equation}
where, $A, B$ are constants, $|X|$ is the number of elements in $X$, and the sum is taken over all the elements.
%%%%%%%%%%%%%

We further assess the classification results with a shallow neural network using the learned representations of the PPMI and ADNI1 datasets. We report standard accuracy and macro averaged precision, recall evaluation metrics,
which are given as: 
\begin{equation}
    Accuracy = \frac{TP+TN}{TP+FP+TN+FN}    
\end{equation}
\begin{equation}
    Recall = \frac{1}{k}\sum_{i=1}^k \frac{TP_i}{TP_i+FN_i}, 
% \end{equation}
% \begin{equation}
    Precision = \frac{1}{k}\sum_{i=1}^k \frac{TP_i}{TP_i+FP_i}
\end{equation}
where, each of TP, TN, FN, FP indicate number of True Positive, True Negative, False Negative and False Positive predictions respectively, and $TP_i$ denotes True Positives for $i^{th}$ class from total $k$ classes.
%%%%%%%%%%%%

\section{Experiments}
\label{sec5}

To evaluate the learning of robust representations through reconstruction, we begin by benchmarking against state-of-the-art methods for 3D reconstruction, 3D VAEs and GANs. GANs are particularly intriguing due to their ability to model complex, high-dimensional data distributions. Among these, $\alpha$-WGAN has demonstrated potential in medical imaging by learning data distributions for tasks such as augmentation. Hence, we compare our R2RG method with the baselines, 3D-VAE and $\alpha$-WGAN.
During training, as discussed in Section \ref{sec42}, we first pre-train each method on the AOMIC dataset, leveraging its larger dataset size. We then use these pre-trained weights and fine-tune them on smaller datasets. 

The comparative experiments on reconstruction are followed by the evaluation of the transferability of the learned representations on a downstream classification task. We compare the representation capabilities of the R2RG model with widely adopted self-supervised baselines, SimCLR and BYOL. Here, similar to our proposed method, we pre-train the baselines SimCLR and BYOL on a larger AOMIC dataset and then fine-tune with the training data from the PPMI and ADNI1 datasets. The train, validation, and test sets are utilized in the downstream classification task using the shallow neural network. A shallow neural network is preferred to rely more on the quality of the learned representations than on the learning capabilities of a deep network. 
For the downstream task, we extract the latent representation vectors from each model and use them as input to a shallow neural network classifier and perform classification on PPMI and ADNI1 datasets.

We experiment further to analyze the representations learned by the R2RG model using the PPMI and ADNI1 datasets. Here, we extract lower dimensional projections of the representation vectors obtained from R2RG, BYOL, and SimCLR, respectively, by applying standard dimensionality reduction techniques. These techniques include Principal Component Analysis (PCA), t-distributed Stochastic Neighbor Embedding (t-SNE), and Uniform Manifold Approximation and Projection for Dimension Reduction (UMAP) \cite{pca,tsne,umap}, which capture both linear and non-linear correlations within the representation vectors. For qualitative analysis, we visualize the projections extracted from these dimensionality reduction methods on a 2D plane. Additionally, we quantify this analysis through a classification task. We choose the PCA components that maintain more than 85\% and 90\% variance in the representation vectors as inputs for a support vector classifier (SVC). 

To further evaluate the effectiveness of our proposed method, we conduct a series of ablation experiments. This involves analyzing the contribution of components of our loss function and discriminator designs on the reconstruction task. In particular, comparisons of the vanilla discriminator and a patch-GAN-based discriminator, as well as the effect of selective removal of $L_1$ and $L_2$ losses from our composite loss. These experiments specifically quantify the practical advantages of the proposed design choices over existing alternatives. Finally, we report the computational efficiency of the proposed method with the inference time required for a single example.

For more details, the implementation of our method and experiments is available at \url{https://anonymous.4open.science/r/3D_MRI_RL/}. The COVID dataset will be made available as per requests.

\section{Results and Discussions}
\label{sec6}
We present our results to answer the following key questions: (i) Is R2RG effective in learning to reconstruct a 3D volume with limited data? If so, how significant is its improvement on the existing methods? (ii) Are the representations learned through reconstruction helpful in downstream classification tasks with even fewer example samples? What can we infer about these learned latents and the latent space? (iii) Does better reconstruction lead to learning better representations? (iv) How does each component of our method contribute to overall learning? Each of the following sections is dedicated to systematically providing experimental evidence on these concerns.

\subsection{Reconstruction}
\label{sec61}
The results of R2RG on reconstruction task are reported in Table \ref{tab:results1} with comparison against baselines.

\begin{table}[h]
    \centering
    \renewcommand{\arraystretch}{1.}
    \caption{Comparison of Mean Structural Similarity Index Measure (MSSIM) and PSNR scores on AOMIC, ADNI1, PPMI, and COVID datasets for different reconstruction methods which avoid using skip connections between encoders and decoders.}
    \begin{tabular}{p{.8cm}
      >{\centering\arraybackslash}p{.6cm}
      >{\centering\arraybackslash}p{.4cm}
      >{\centering\arraybackslash}p{.6cm}
      >{\centering\arraybackslash}p{.4cm}
      >{\centering\arraybackslash}p{.6cm}
      >{\centering\arraybackslash}p{.4cm}
      >{\centering\arraybackslash}p{.6cm}
      >{\centering\arraybackslash}p{.6cm}
    }
    % \toprule
    \hline
        Method & \multicolumn{2}{c}{AOMIC} & \multicolumn{2}{c}{ADNI1} & \multicolumn{2}{c}{COVID} & \multicolumn{2}{c}{PPMI} \\
         & MSSIM & PSNR & MSSIM & PSNR & MSSIM & PSNR & MSSIM & PSNR \\
         \hline
         $\alpha$-WGAN & 0.707 & 20.33 & 0.589 & 19.27 & 0.673 & 18.98 & 0.592 & 18.99 \\
         VAE & 0.898 & 22.48 & 0.895 & 25.75 & 0.909 & 24.01 & 0.807 & 22.56 \\
         R2RG & \textbf{0.947} & \textbf{29.11} & \textbf{0.921} & \textbf{27.14} & \textbf{0.944} & \textbf{27.57} & \textbf{0.834} & \textbf{23.24} \\
         \hline
    \end{tabular}
    \label{tab:results1}
\end{table}

With quantitative evaluation, our method achieves mean SSIM scores exceeding 0.92 on both the ADNI1 and COVID datasets, and establishing a reconstruction benchmark on the AOMIC dataset with an MSSIM of 0.94, while also attaining a superior MSSIM score of 0.834 on the challenging PPMI dataset. The results are consistent through both the SSIM and PSNR evaluations. The PPMI dataset exhibits inherent aleatoric uncertainty (arising from its heterogeneous data acquisition) that makes learning particularly difficult, given that it contains a comparatively smaller number of data. Nonetheless, R2RG maintains outperforming baselines, demonstrating generalizability on a varying yet smaller dataset. A general trend of better results on the AOMIC dataset in all the methods can be observed in Table \ref{tab:results1}. These outcomes can be credited to its larger dataset size. 

The 3D $\alpha$-WGAN baseline employs a code discriminator to enforce a Gaussian prior in the latent space, which requires its generator to reconstruct real volumes from sampled random noise. However, the task of reconstructing from random noise proves highly challenging and limits the performance of the model. The 3D VAE \cite{vae3d}, in contrast, incorporates a distributional shift penalty to enforce its latent space towards a Gaussian distribution. However, it relies exclusively on the encoded representation vector for reconstructing the real volume, resulting in better performance, but suffers from averaging in the dataset and consequently produces blurry outputs \cite{neur2}. These findings suggest that (i) the code-discriminator component used in $\alpha$-WGAN, which enforces reconstruction from random noise, limits its quantitative reconstruction capabilities, (ii) 3D VAE, with only a focus on the reconstruction task, though it achieves the closest comparable results, produces over-smoothed outputs. This can be attributed to the fact that the 3D VAE loss does not explicitly incentivize learning the distribution of the volume space. Overall, R2RG consistently outperforms other methods in the reconstruction task on all four datasets, demonstrating superior ability to transfer knowledge on a smaller and new COVID dataset, and to generalize well on the diverse PPMI dataset.

\begin{figure*}[h]
    \centering
    \renewcommand{\arraystretch}{2.36}
    \begin{tabular}{m{2.4cm}>{\centering\arraybackslash}p{3.31cm}>{\centering\arraybackslash}p{3.31cm}>{\centering\arraybackslash}p{3.31cm}}
        \textbf{Method} & \textbf{AOMIC} & \textbf{ADNI1} & \textbf{COVID}\\
        Real & \adjustimage{valign=c,width=3.31cm}{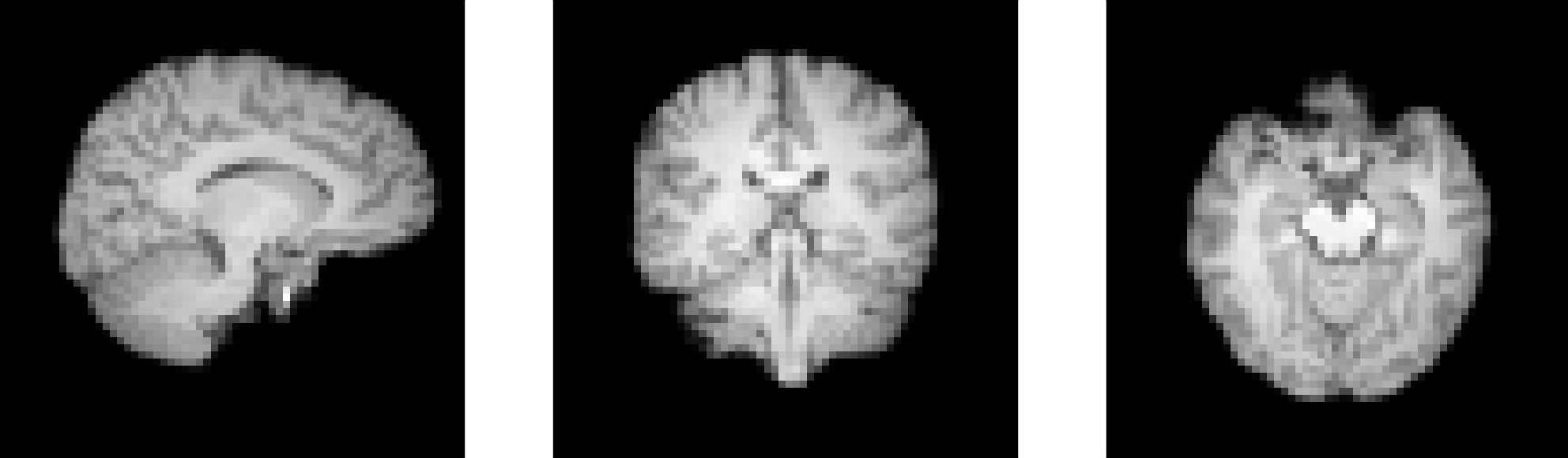} & \adjustimage{valign=c, width=3.31cm}{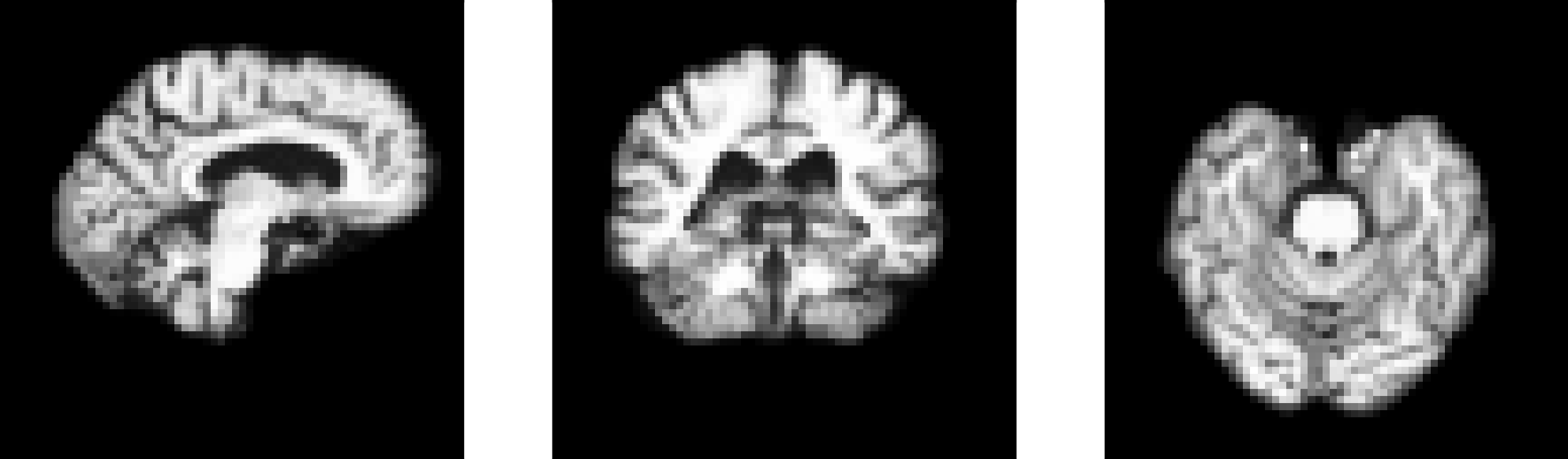} & \adjustimage{valign=c, width=3.31cm}{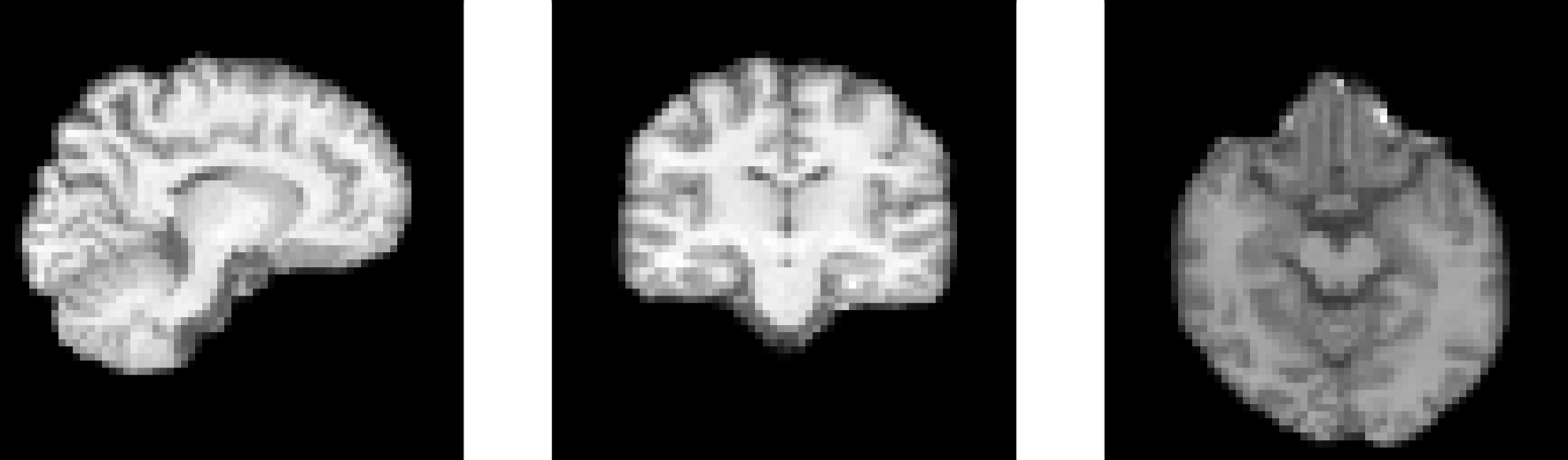}\\
        $\alpha$-WGAN  & \adjustimage{valign=c, width=3.31cm}{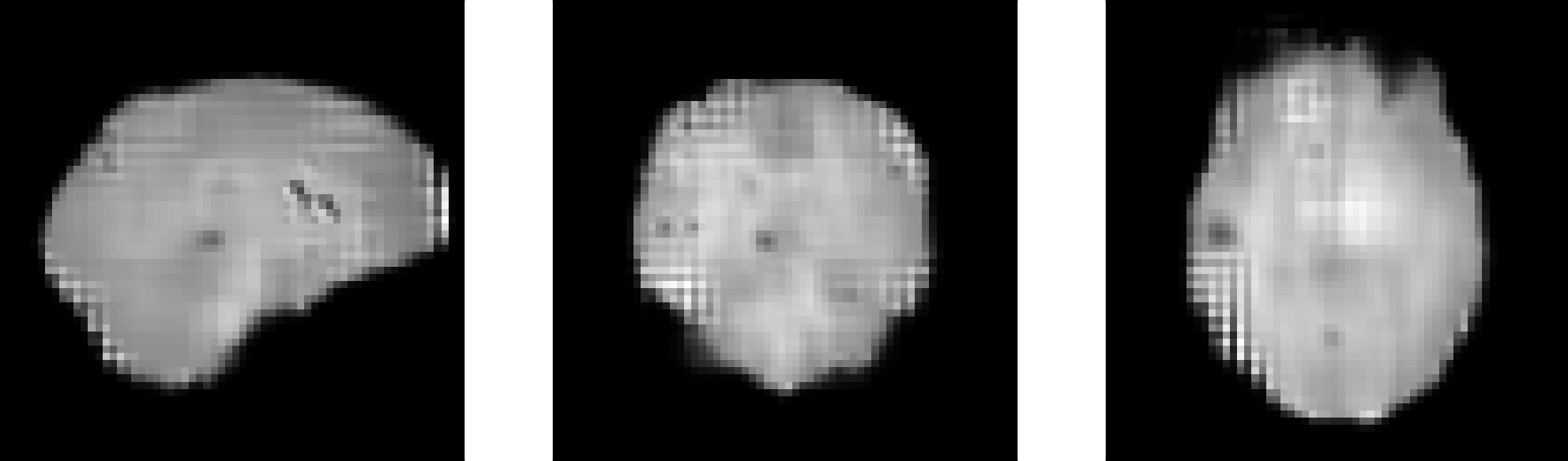} & \adjustimage{valign=c, width=3.31cm}{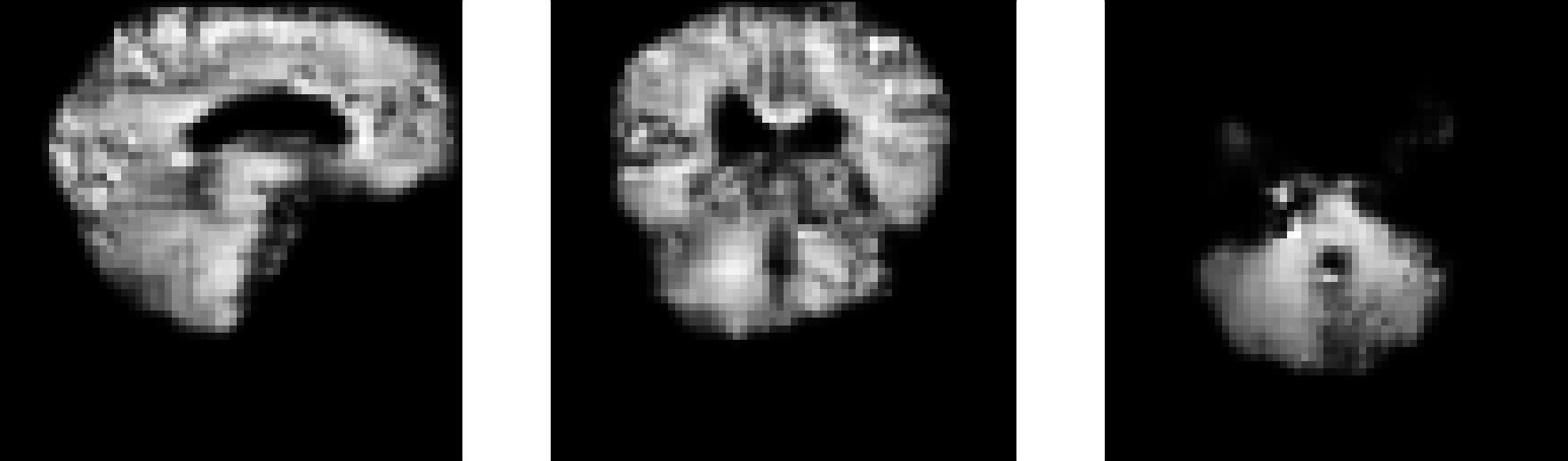} & \adjustimage{valign=c,width=3.31cm}{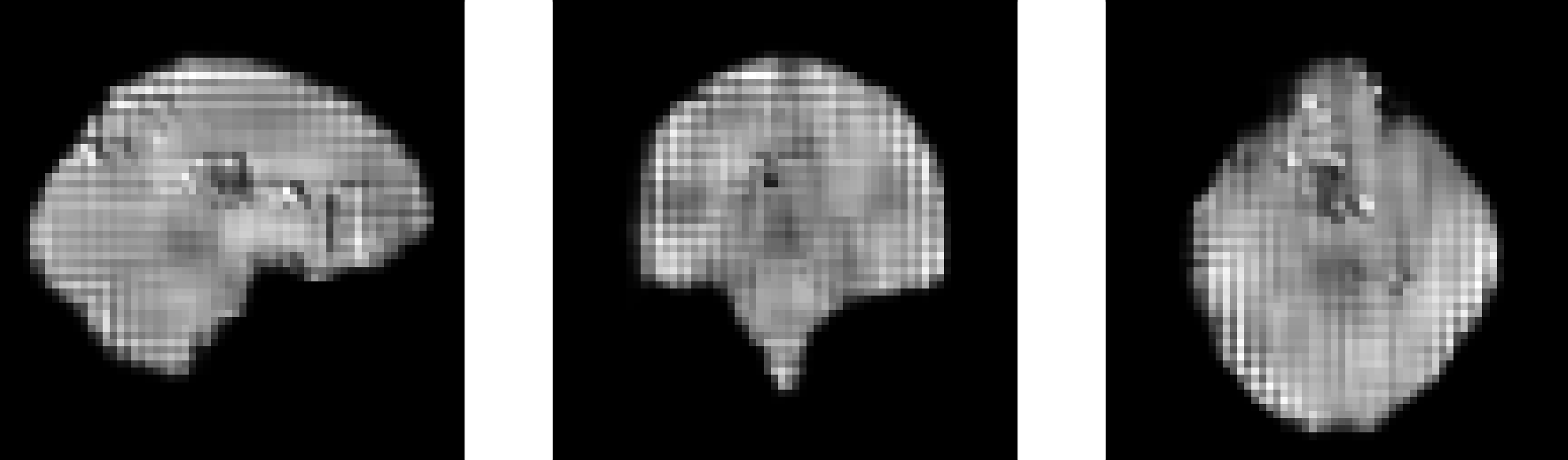} \\
        VAE  & \adjustimage{valign=c,width=3.31cm}{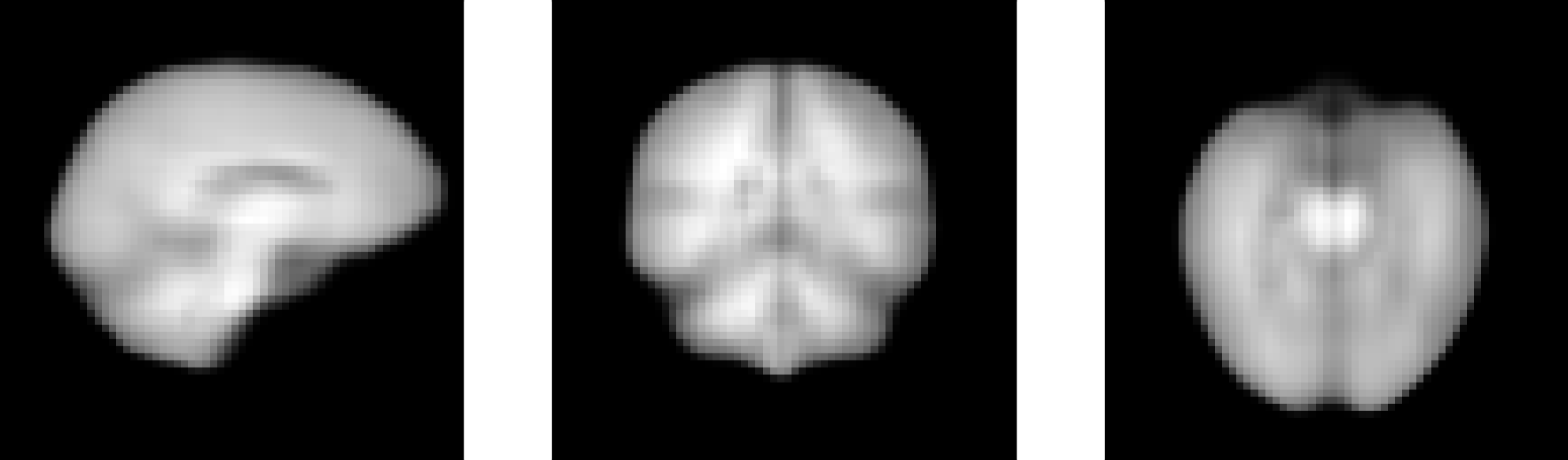} & \adjustimage{valign=c,width=3.31cm}{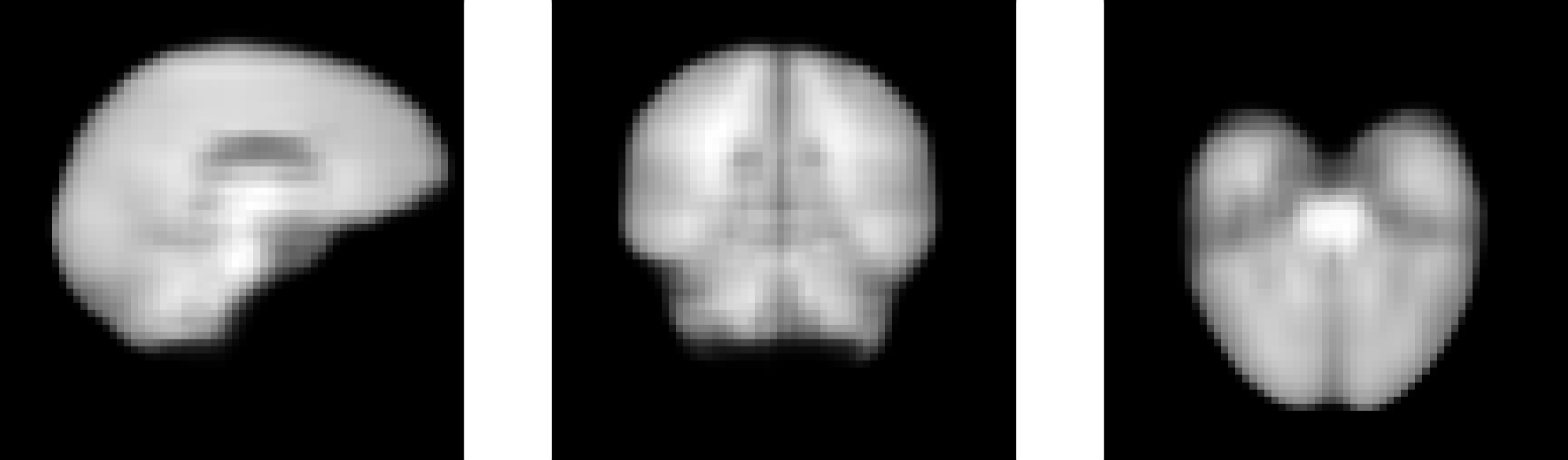} & \adjustimage{valign=c, width=3.31cm}{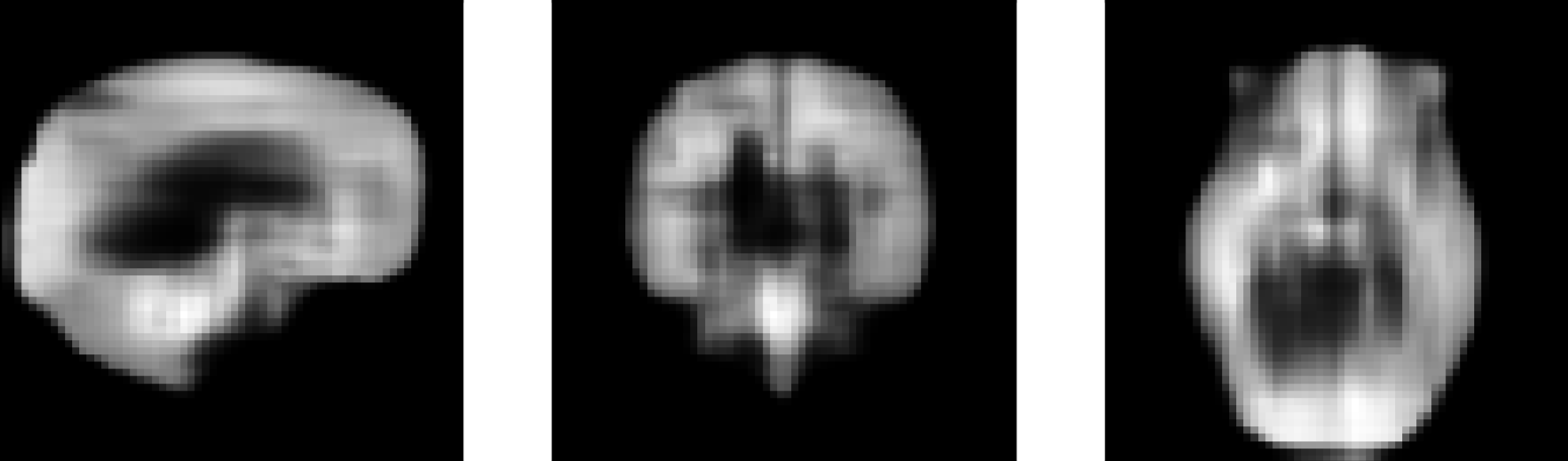} \\
        \footnotesize{R2RG (only $L_1$)} & \adjustimage{valign=c,width=3.31cm}{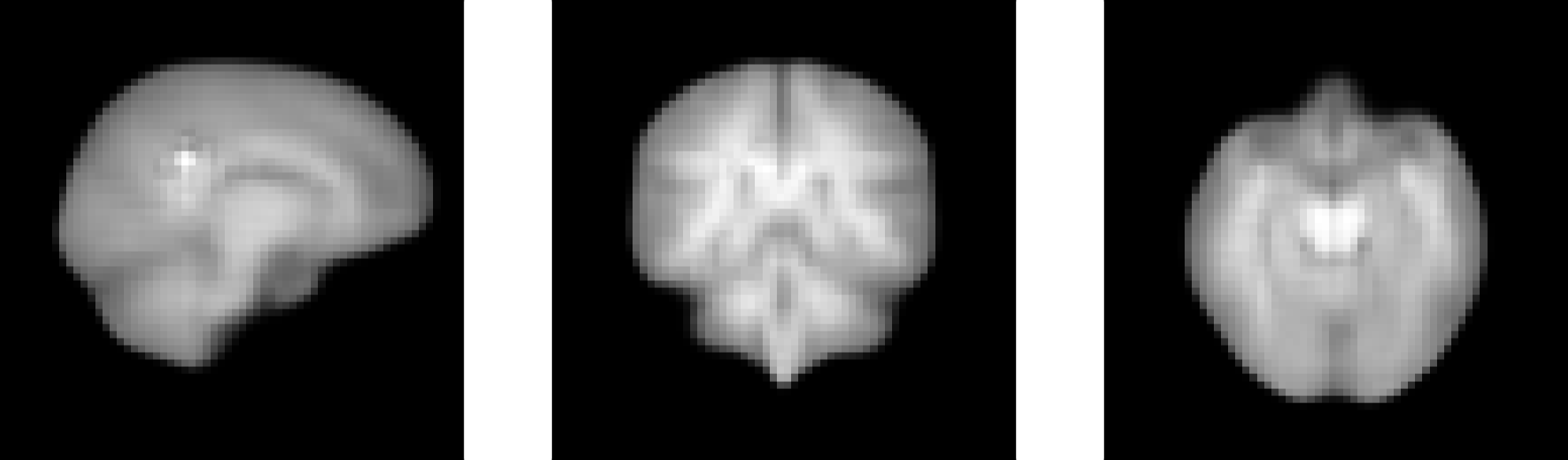} & \adjustimage{valign=c,width=3.31cm}{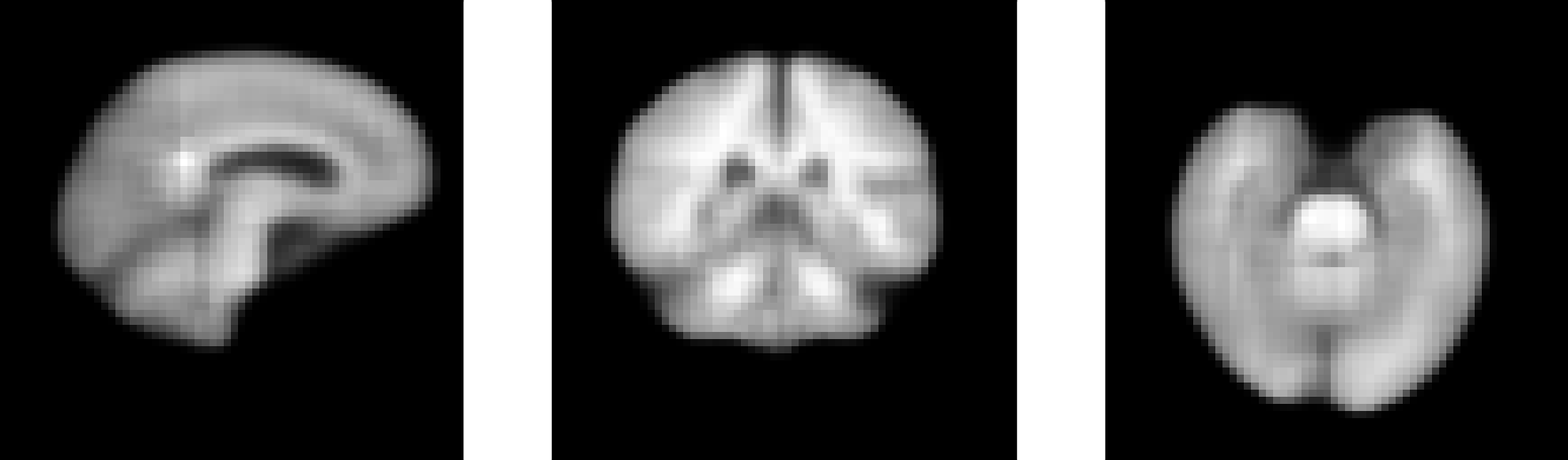} & \adjustimage{valign=c,width=3.31cm}{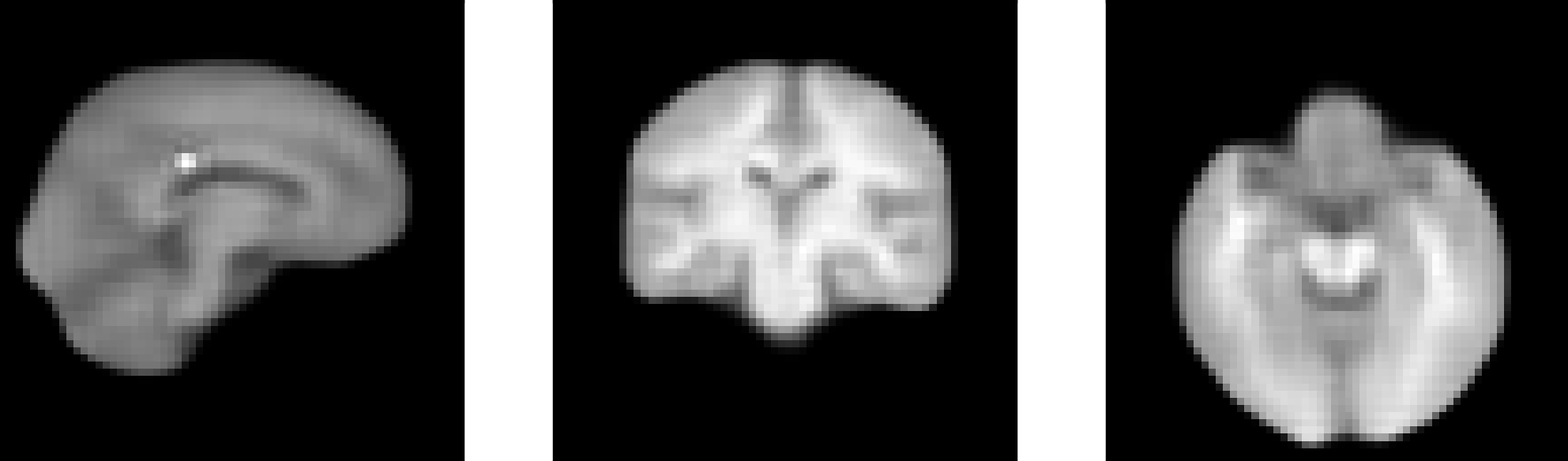} \\
        \footnotesize{R2RG} & \adjustimage{valign=c,width=3.31cm}{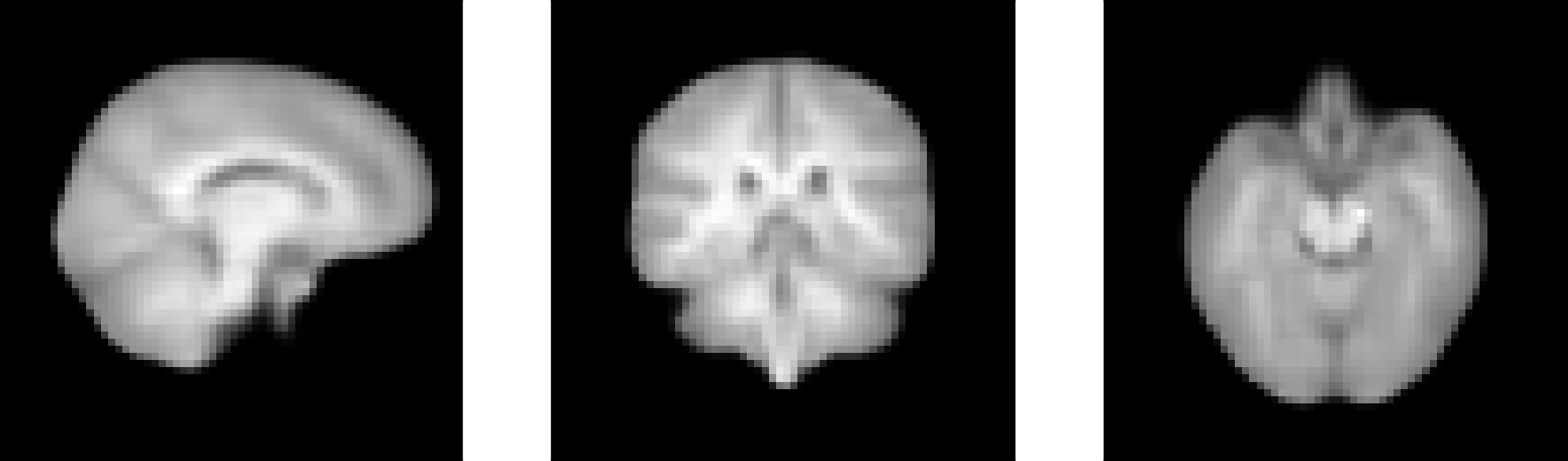} & \adjustimage{valign=c,width=3.31cm}{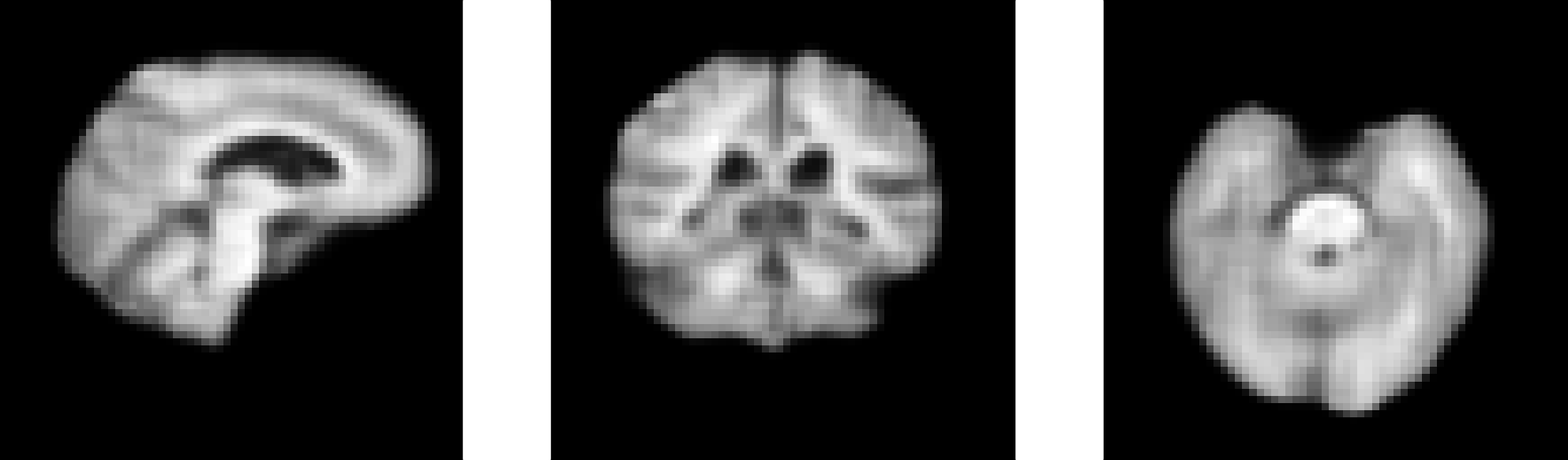} & \adjustimage{valign=c,width=3.31cm}{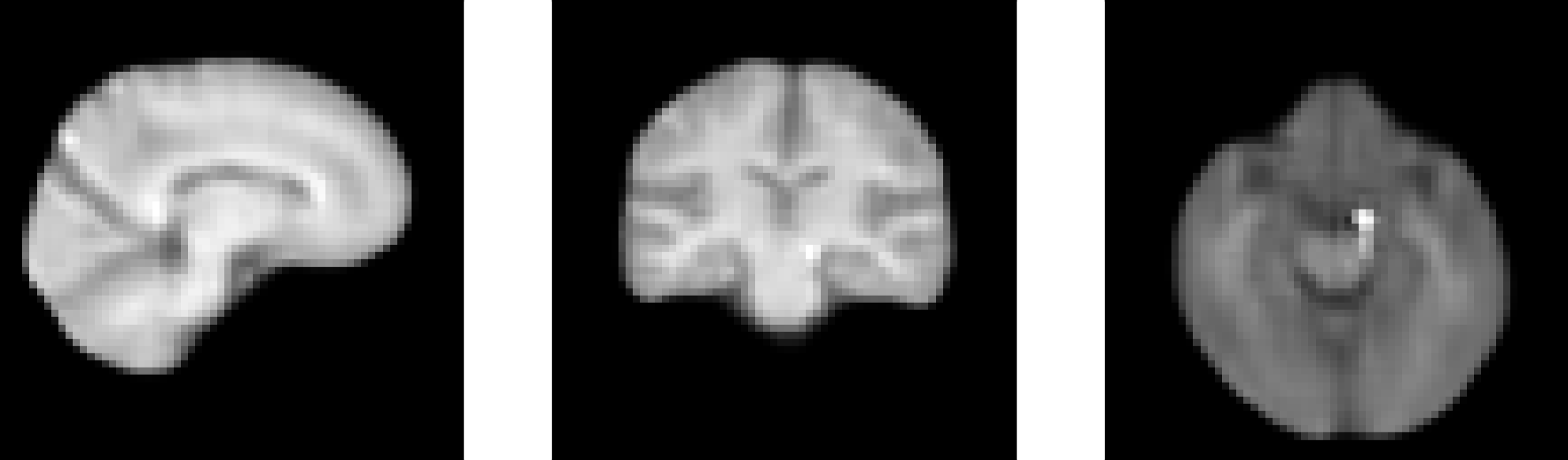} \\
    \end{tabular}
    \caption{Qualitative results of brain volume reconstruction across different methods: each row represents results on a method named in the left column; three columns from each dataset include sagittal, coronal, and axial views of the volume from each method, respectively.}
    \label{fig:out_imgs}
\end{figure*}

The qualitative results in Fig. \ref{fig:out_imgs}, with slices from orthogonal views, show similar outcomes. Our method reconstructs images with higher fidelity and greater structural and luminance similarity than baselines, with composite loss showing slight improvements over the blurry effect observed when the $L_2$ loss is omitted. In contrast, 3D VAE exhibits significant blurriness and poor generalization on the smaller COVID dataset, while $\alpha$-WGAN, despite capturing structural outlines more effectively with its additional discriminator, fails to generate high-fidelity volumes.

In summary, these results show that R2RG not only outperforms existing approaches in terms of reconstruction quality, but also generalizes more effectively across various datasets.

\subsection{Classification}% Evaluations}
\label{sec63}

In the aforementioned results, we demonstrate that the representations learned by our proposed method yield superior reconstruction performance compared to existing 3D reconstruction algorithms that avoid information leakage through skip connections.
To further assess the utility and generalizability of these representations, we conduct a downstream classification task using the PPMI and ADNI1 datasets. In addition, we compare the performance of our approach against latent representations generated by two widely adopted self-supervised learning methods SimCLR, BYOL, as well as with representations learned from the reconstruction baselines VAE, $\alpha$-WGAN.

As reported in Table \ref{tab:results3}, on the smaller PPMI dataset, the representations learned by R2RG yield a substantial performance boost over SimCLR and BYOL, achieving at least a 10\% improvement in both classification accuracy and precision, along with a 3\% increase in recall. These results underscore the effectiveness of R2RG in capturing clinically relevant features that generalize well to diagnostic tasks, highlighting structural fidelity of the latent space. On the ADNI1 dataset, with increasing dataset size, BYOL performs closer to R2RG. R2RG also improves in performance with a larger dataset and outperforms other methods. These results demonstrate the consistency of our method and also validate the hypothesis that better reconstruction results in richer representations and, consequently, better downstream accuracy. Moreover, it is interesting to note that reconstruction with GANs results in better representations for downstream tasks than a simple encoding-decoding architecture.

\begin{table}[h]
    \centering
    \renewcommand{\arraystretch}{1.15}
    \caption{Classification results on learned representations of the PPMI and ADNI1 datasets through representation learning methods with a shallow Neural Network classifier.}
    \begin{tabular}{m{1.2cm}>{\centering\arraybackslash}p{.85cm}>{\centering\arraybackslash}p{.85cm}>{\centering\arraybackslash}p{.6cm}>{\centering\arraybackslash}p{.85cm}>{\centering\arraybackslash}p{.85cm}>{\centering\arraybackslash}p{.6cm}}
    \hline
         Method & \multicolumn{3}{c}{PPMI} & \multicolumn{3}{c}{ADNI1} \\
          & Accuracy & Precision & Recall & Accuracy & Precision & Recall \\
         \hline
         SimCLR & 0.581 & 0.564 & 0.659 & 0.510 & 0.345 & 0.416 \\
         BYOL & 0.621 & 0.603 & 0.727 & 0.848 & {0.828} & \textbf{0.831} \\
         VAE & 0.609 & 0.607 & 0.681 & 0.759 & 0.792 & 0.735 \\
         $\alpha$-WGAN & 0.655 & 0.640 & 0.727 & 0.807 & 0.806 & 0.773 \\
         R2RG & \textbf{0.712} & \textbf{0.702} & \textbf{0.750} & \textbf{0.855} & \textbf{0.843} & {0.825} \\
         \hline
    \end{tabular}
    \label{tab:results3}
\end{table}

\begin{table}[h]
    \centering
    \renewcommand{\arraystretch}{1.15}
    \caption{Classification results from a support vector classifier (SVC) with principal components maintaining variance \textit{p} of learned representations of the PPMI and ADNI1 datasets.}
    \begin{tabular}{m{.05cm}m{1.2cm}>{\centering\arraybackslash}p{.8cm}>{\centering\arraybackslash}p{.7cm}>{\centering\arraybackslash}p{.6cm}>{\centering\arraybackslash}p{.8cm}>{\centering\arraybackslash}p{.7cm}>{\centering\arraybackslash}p{.6cm}}
    \hline
         \textit{p} & Method & \multicolumn{3}{c}{PPMI} & \multicolumn{3}{c}{ADNI1}\\
          & & Accuracy & Precision & Recall & Accuracy & Precision & Recall \\
         \hline
         \multirow{3}{*}{\rotatebox{90}{0.85}} & SimCLR & 0.442 & 0.461 & 0.545 & 0.483 & 0.433 & 0.403 \\
                            & BYOL & {0.574} & {0.568} & {0.659} & 0.469 & 0.493 & 0.422 \\
                            & VAE & 0.540 & 0.543 & 0.568 & 0.393 & 0.359 & 0.355 \\
                            & $\alpha$-WGAN & 0.597 & 0.576 & 0.697 & 0.413 & 0.405 & 0.380 \\
                            & R2RG & 0.529 & 0.528 & 0.636 & 0.351 & 0.303 & 0.308 \\
         \hline
         \multirow{3}{*}{\rotatebox{90}{0.90}} & SimCLR & 0.407 & 0.431 & 0.50 & 0.496 & 0.495 & 0.414 \\
                            & BYOL & {0.597} & {0.591} & {0.659} & 0.413 & 0.353 & 0.361 \\
                            & VAE & 0.551 & 0.551 & 0.613 & 0.352 & 0.312 & 0.310 \\
                            & $\alpha$-WGAN & 0.609 & 0.586 & 0.668 & 0.427 & 0.397 & 0.381 \\
                            & R2RG & 0.528 & 0.529 & 0.613 & 0.406 & 0.346 & 0.354 \\
         \hline
    \end{tabular}
    \label{tab:results4}
\end{table}

\begin{figure*}[!t]
    \centering
    \renewcommand{\arraystretch}{2.36}
    \begin{tabular}{m{1.4cm}>{\centering\arraybackslash}p{3.41cm}>{\centering\arraybackslash}p{3.41cm}>{\centering\arraybackslash}p{3.41cm}}
        {Method} & {PCA} & {t-SNE} & {UMAP}\\
        % \multicolumn{4}{l}{\small\textit{PPMI Dataset}} \\[-2ex]
        \small{R2RG} & \adjustimage{trim=20mm 15mm 20mm 20mm,clip,valign=c,width=3.41cm}{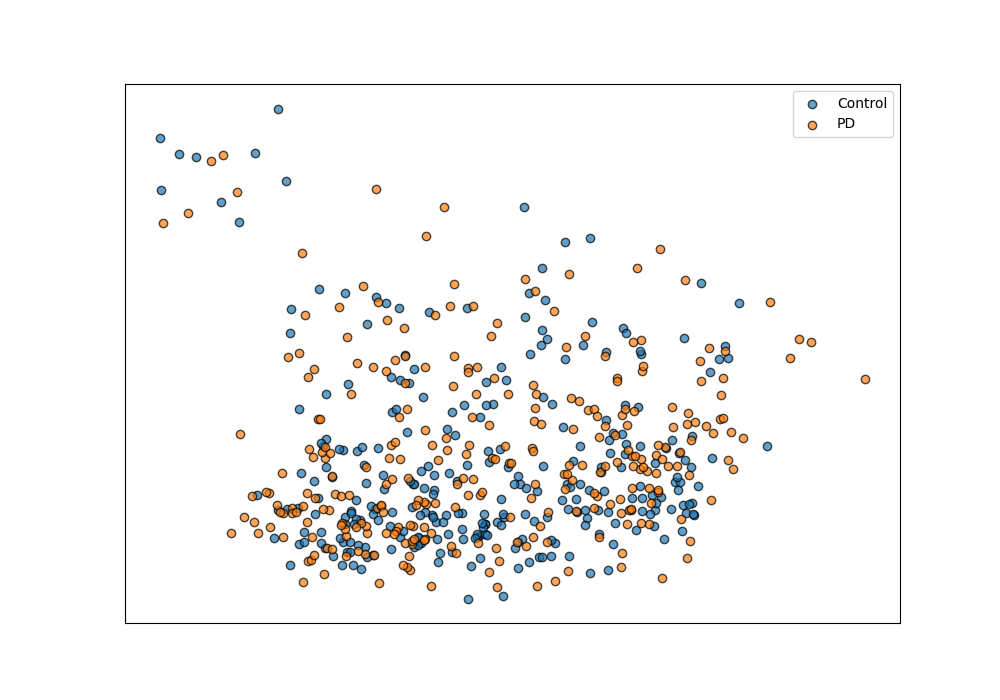} & \adjustimage{trim=20mm 15mm 20mm 20mm,clip,valign=c,width=3.41cm}{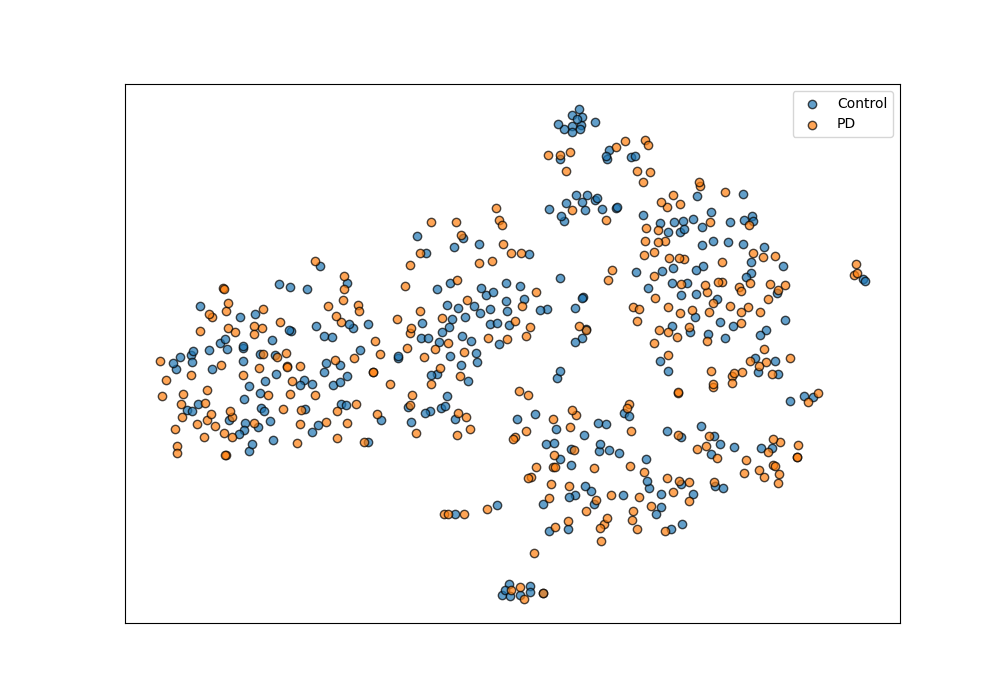} & \adjustimage{trim=20mm 15mm 20mm 20mm,clip,valign=c,width=3.41cm}{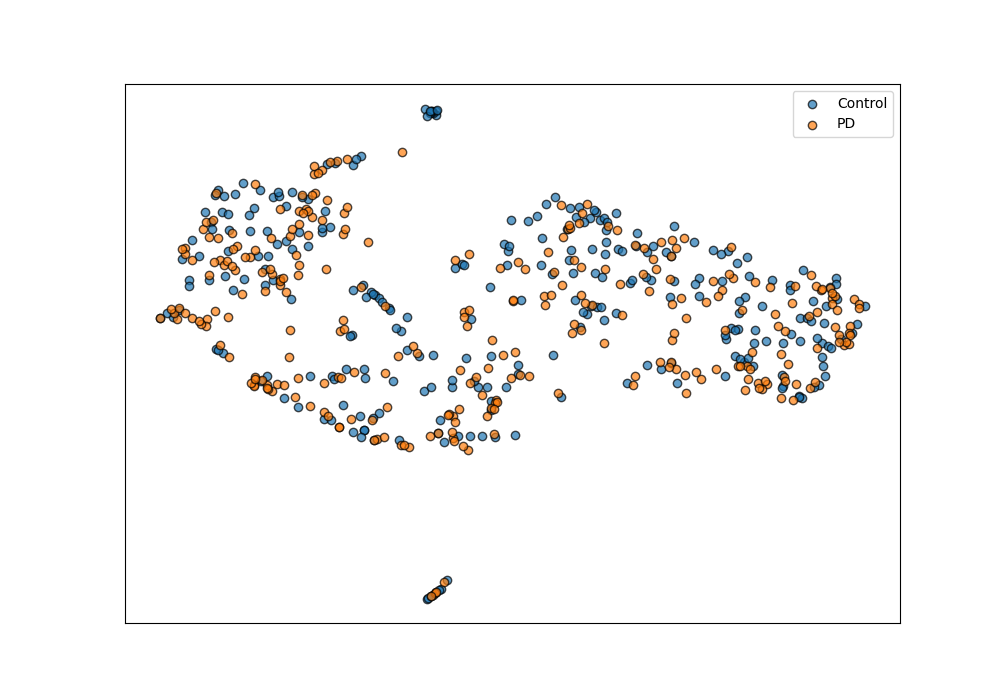}\\
        \small{BYOL} & \adjustimage{trim=20mm 15mm 20mm 20mm,clip,valign=c,width=3.41cm}{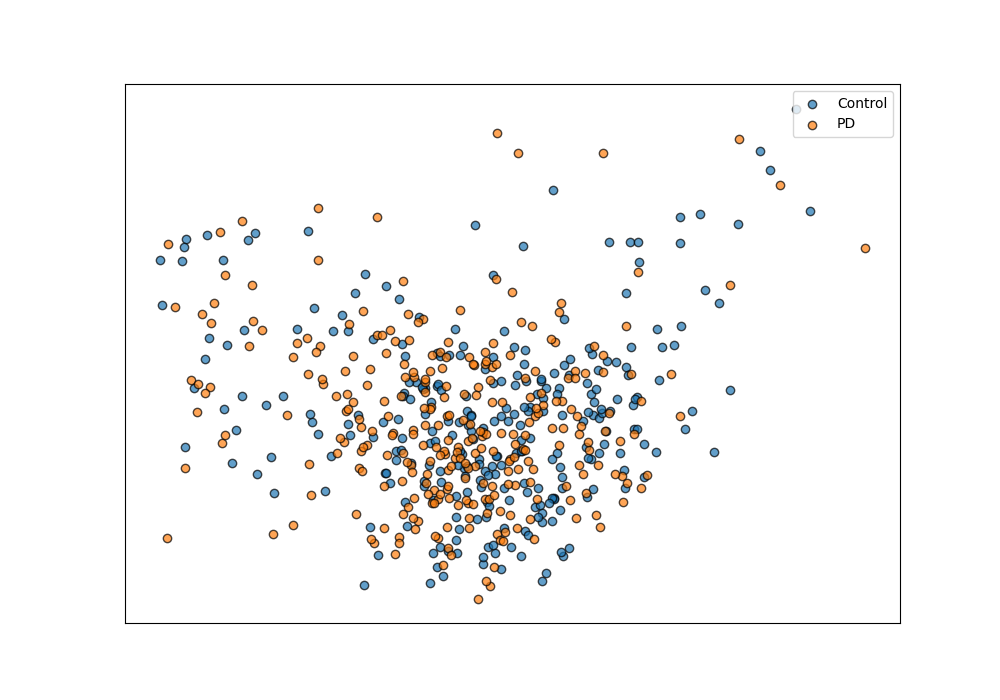} & \adjustimage{trim=20mm 15mm 20mm 20mm,clip,valign=c,width=3.41cm}{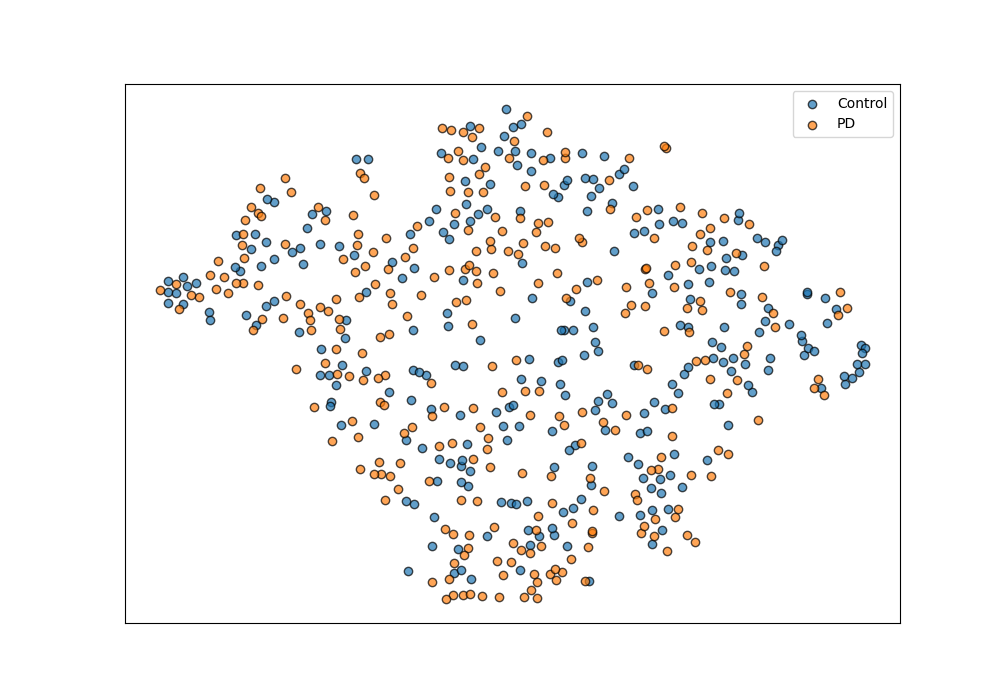} & \adjustimage{trim=20mm 15mm 20mm 20mm,clip,valign=c,width=3.41cm}{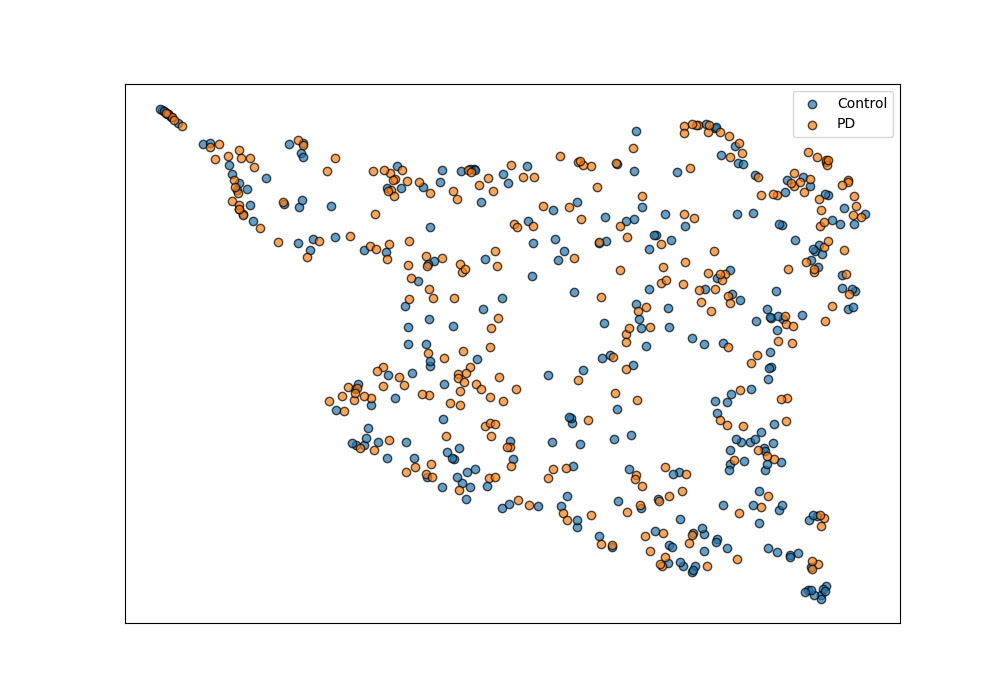}\\
        \small{SimCLR} & \adjustimage{trim=20mm 15mm 20mm 20mm,clip,valign=c,width=3.41cm}{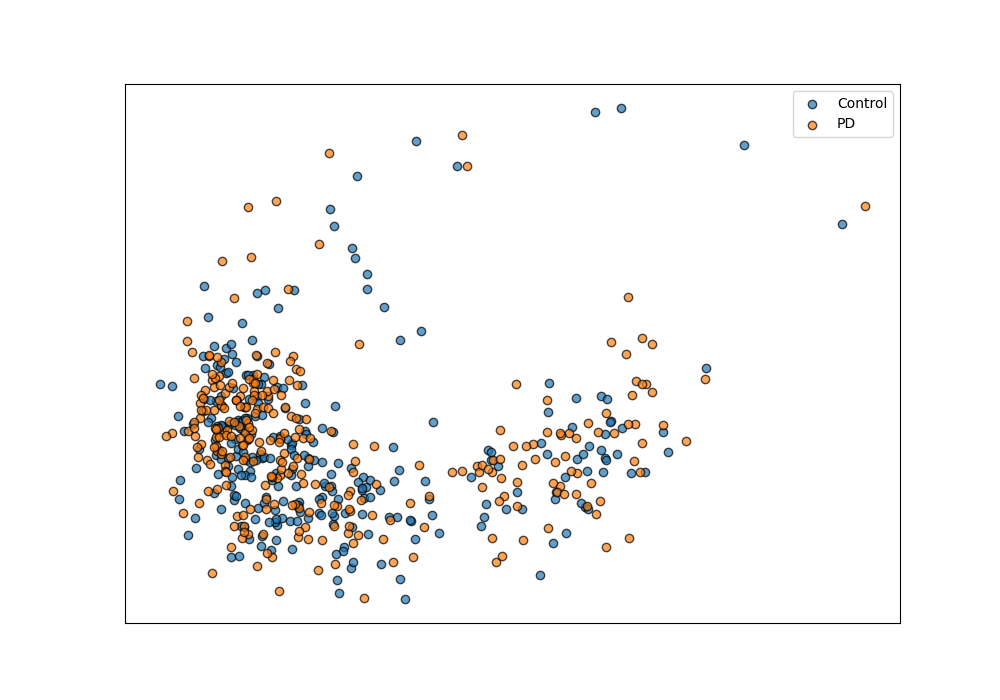} & \adjustimage{trim=20mm 15mm 20mm 20mm,clip,valign=c,width=3.41cm}{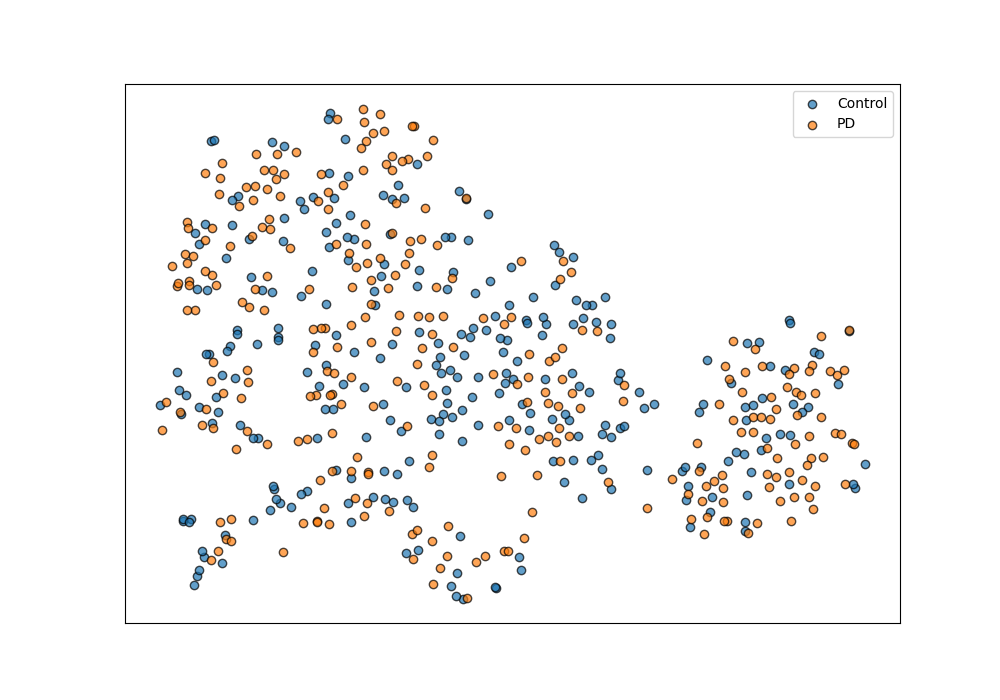} & \adjustimage{trim=20mm 15mm 20mm 20mm,clip,valign=c,width=3.41cm}{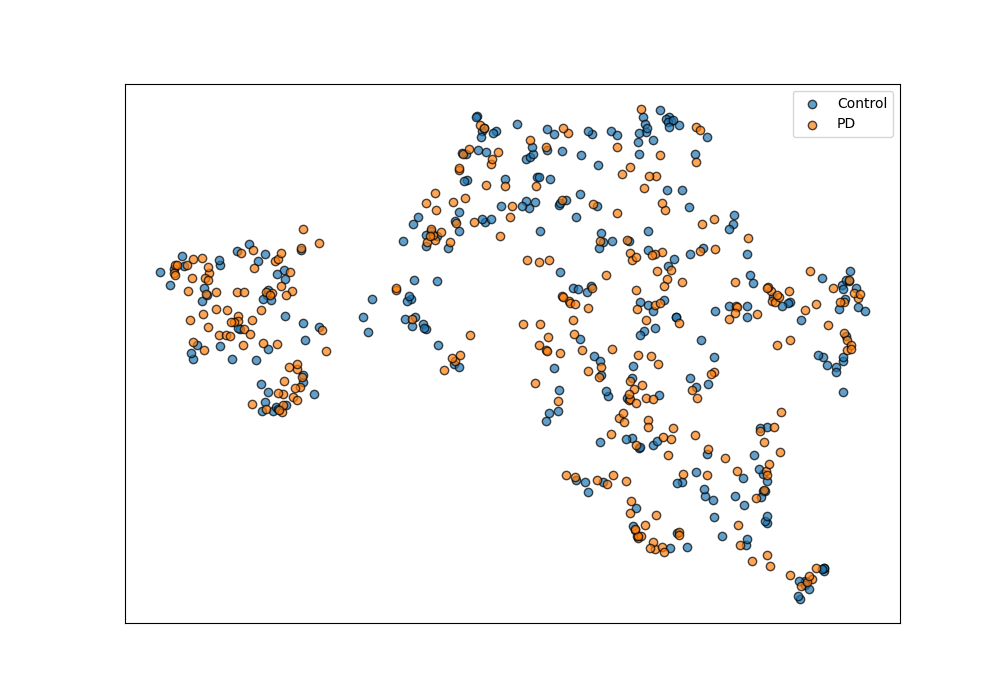}\\
        \small{R2RG} & \adjustimage{trim=20mm 15mm 20mm 20mm,clip,valign=c,width=3.41cm}{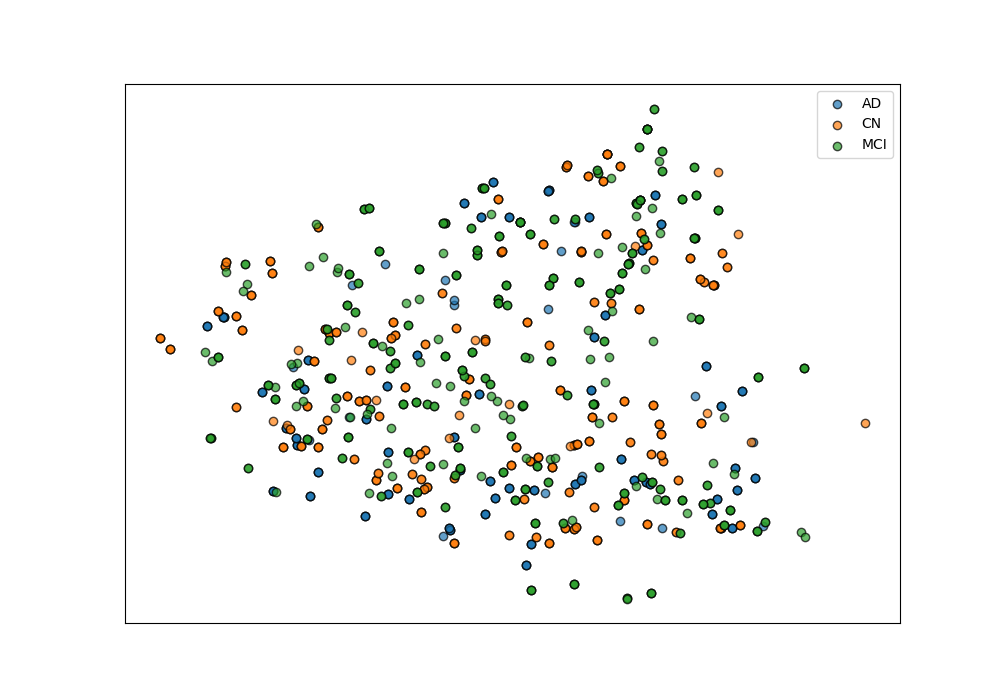} & \adjustimage{trim=20mm 15mm 20mm 20mm,clip,valign=c,width=3.41cm}{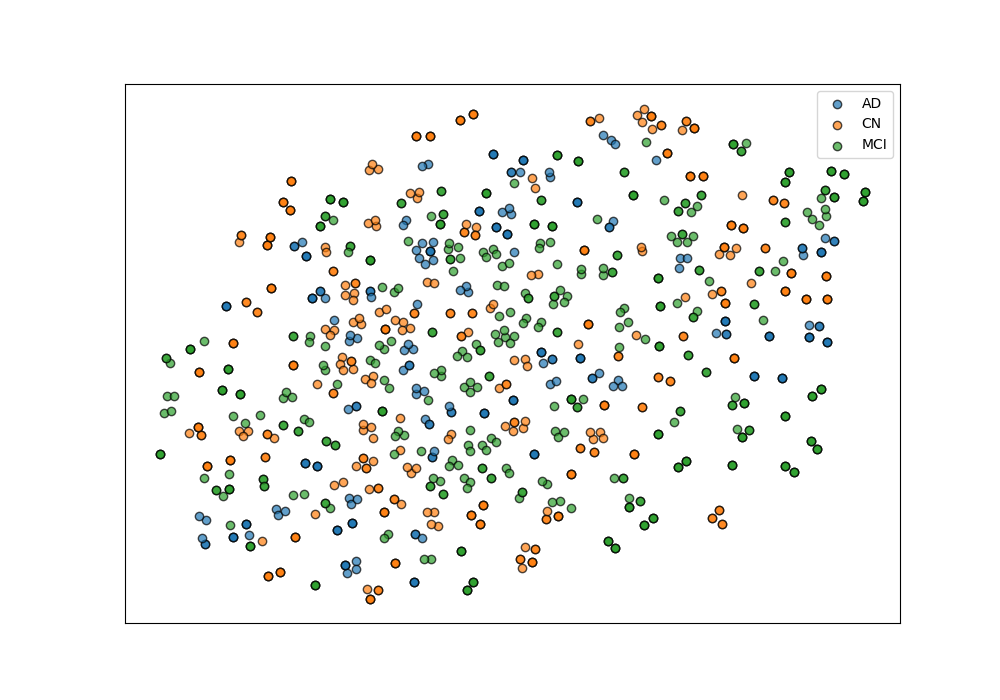} & \adjustimage{trim=20mm 15mm 20mm 20mm,clip,valign=c,width=3.41cm}{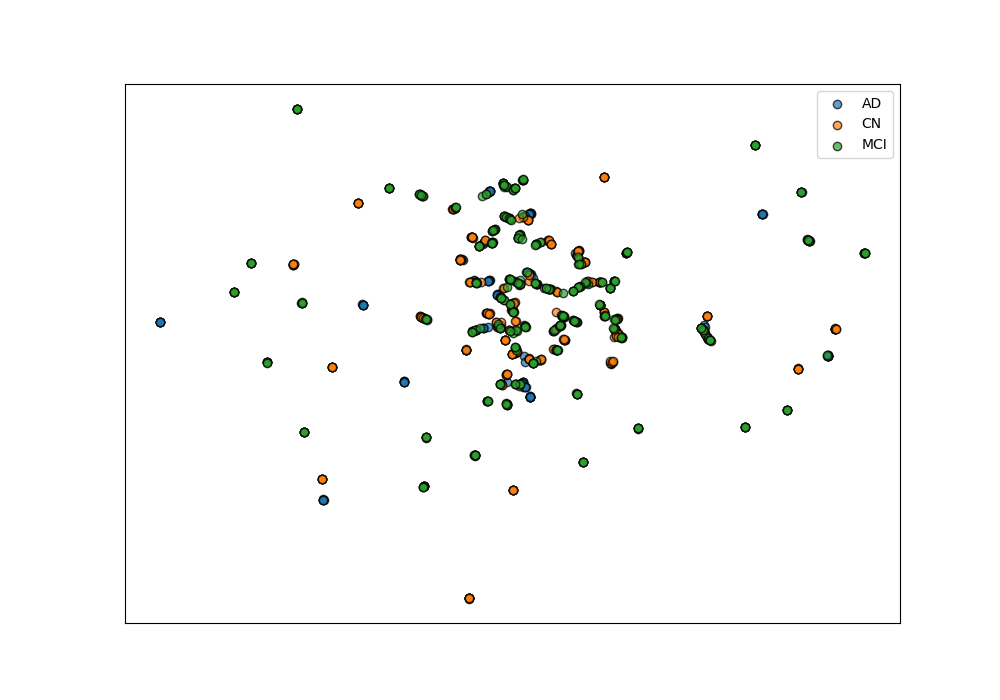}\\
        \small{BYOL} & \adjustimage{trim=20mm 15mm 20mm 20mm,clip,valign=c,width=3.41cm}{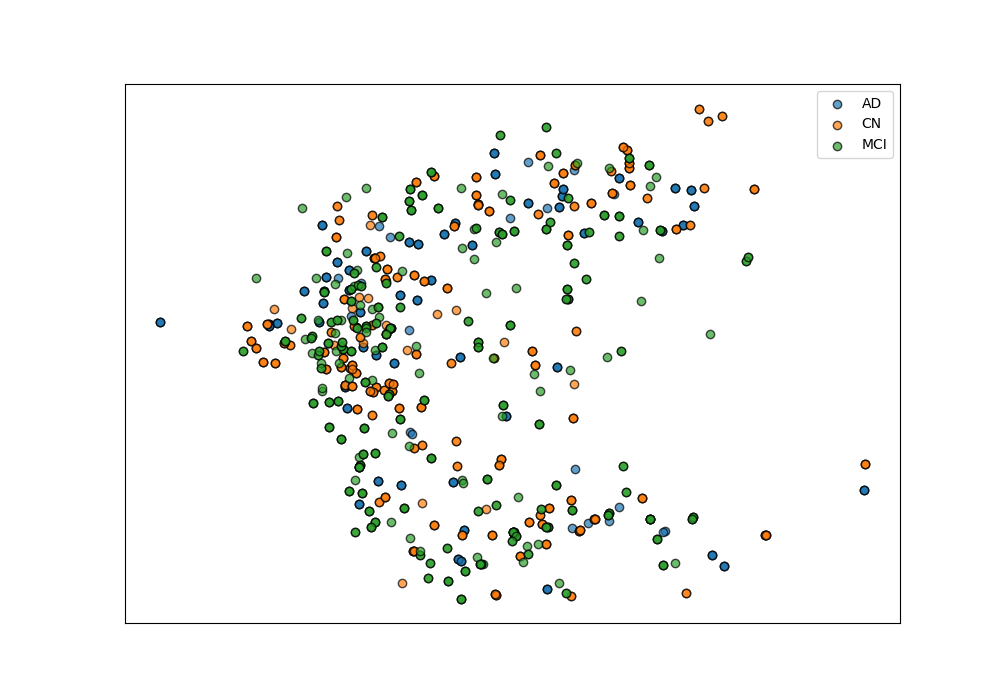} & \adjustimage{trim=20mm 15mm 20mm 20mm,clip,valign=c,width=3.41cm}{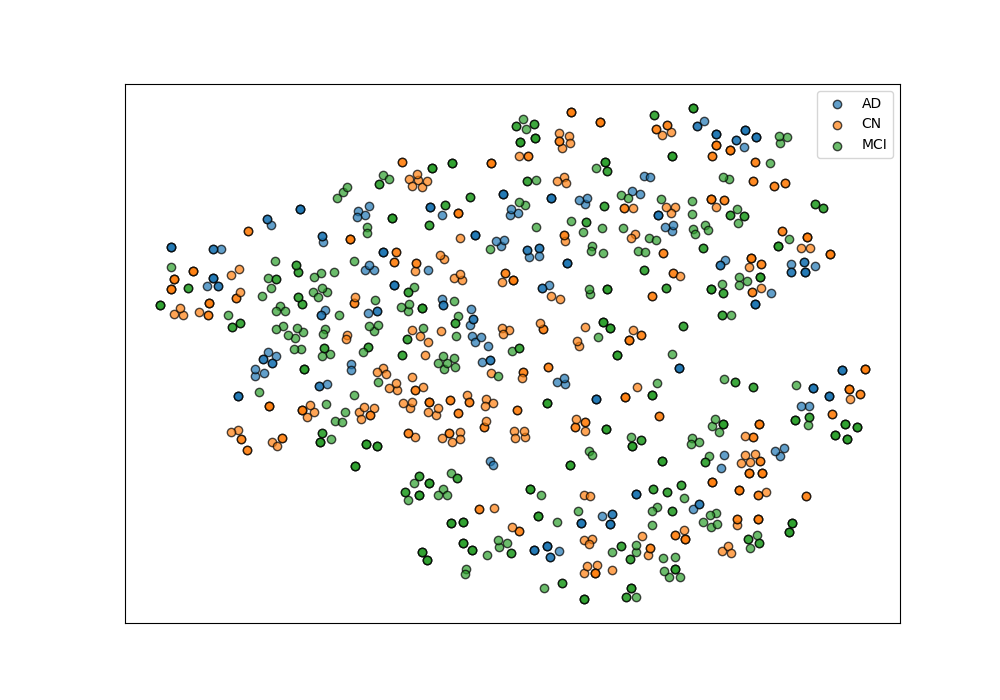} & \adjustimage{trim=20mm 15mm 20mm 20mm,clip,valign=c,width=3.41cm}{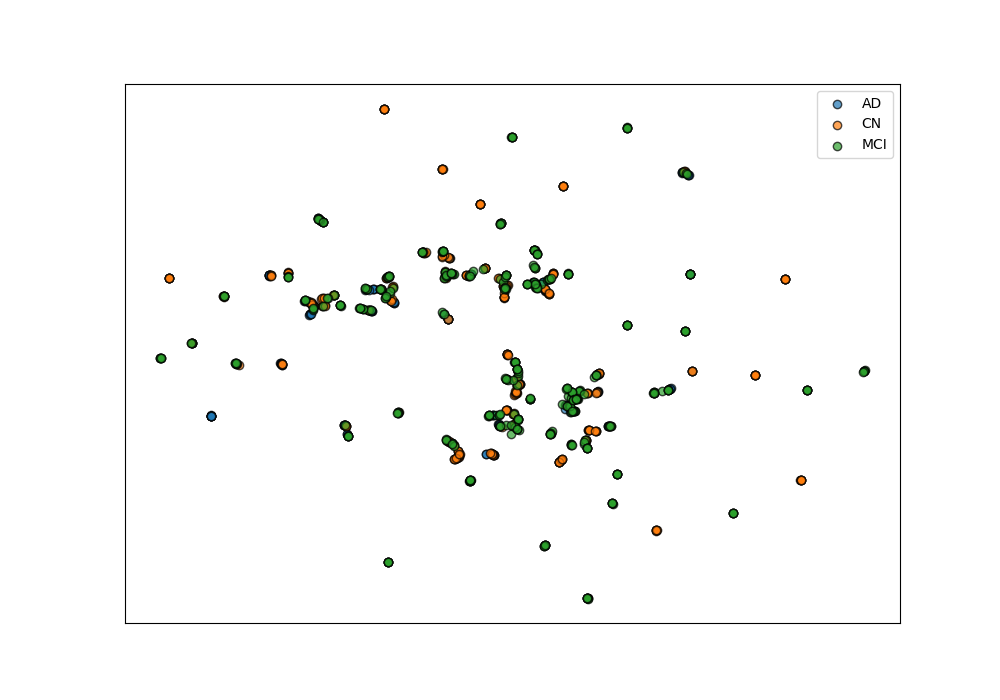}\\
        \small{SimCLR} & \adjustimage{trim=20mm 15mm 20mm 20mm,clip,valign=c,width=3.41cm}{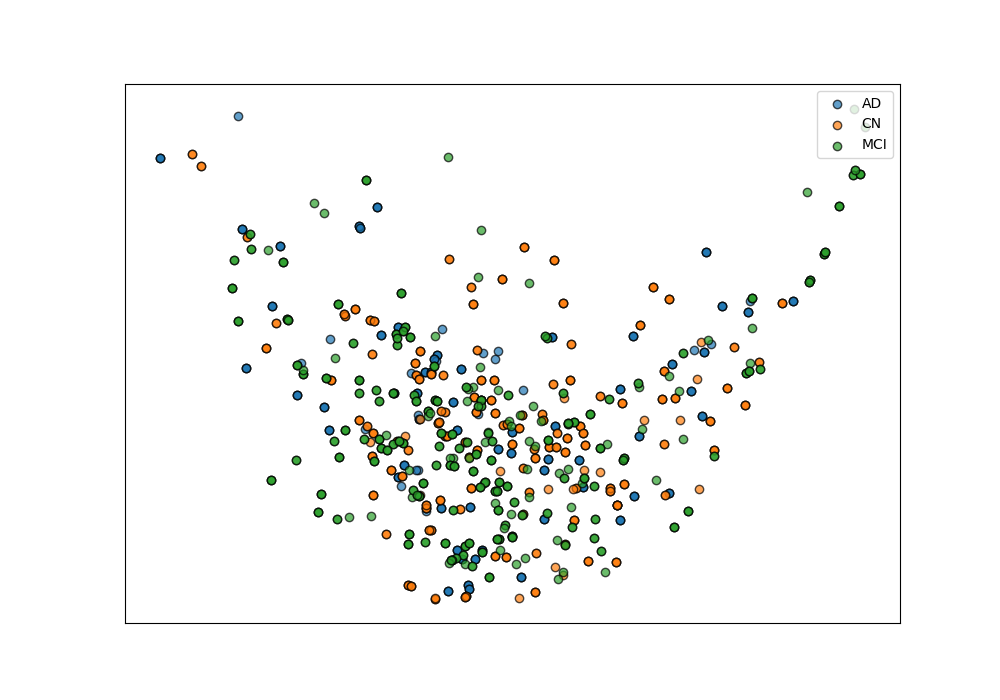} & \adjustimage{trim=20mm 15mm 20mm 20mm,clip,valign=c,width=3.41cm}{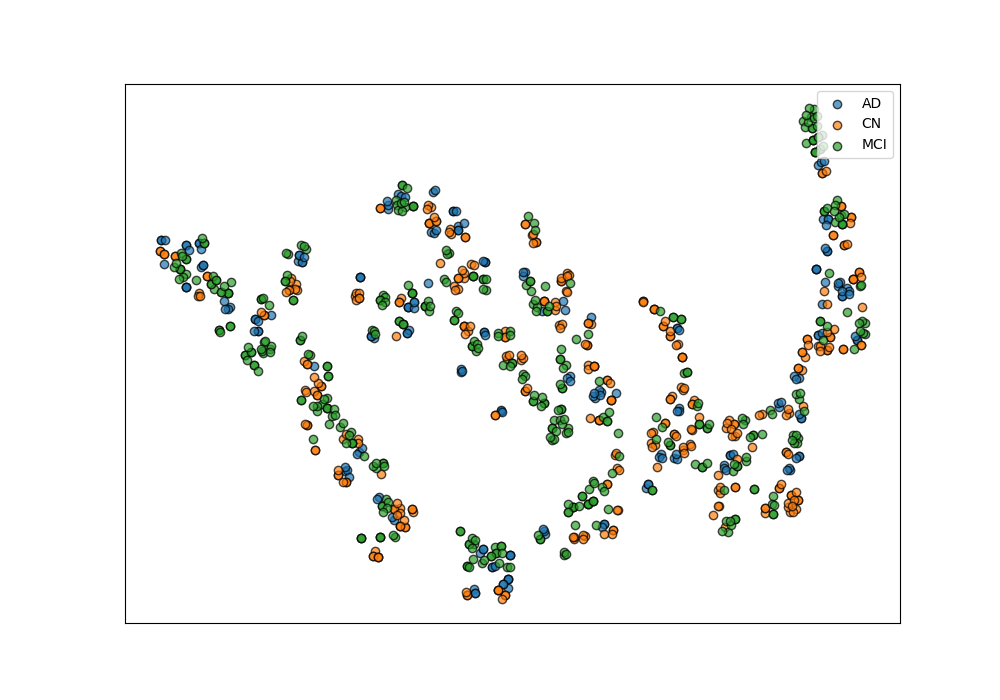} & \adjustimage{trim=20mm 15mm 20mm 20mm,clip,valign=c,width=3.41cm}{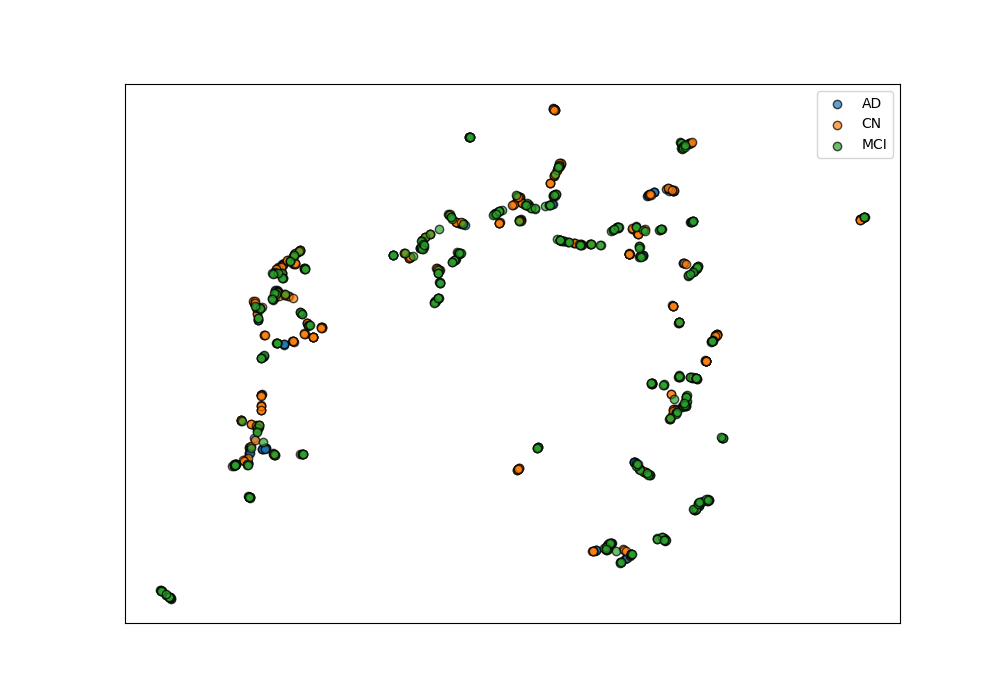}\\
    \end{tabular}
    \caption{Visualizations of the projections extracted by PCA, t-SNE, and UMAP from representation vectors learned by R2RG, BYOL, and SimCLR. First three rows show visualizations from the PPMI dataset (Control: blue, PD: orange) and the latter three on the ADNI1 dataset (AD: blue, CN: orange, MCI: green). It can be observed that these projections are not separable, although it should be noted that R2RG representation are spread more widely.}
    \label{fig:comps}
\end{figure*}

We attribute this improvement to the architectural design of our approach, which is explicitly modeled to encode latent representations such that they preserve spatially rich features required to reconstruct the full 3D input volume. Our composite loss forces the encoder to capture fine-grained anatomical details and discriminator provides the volumetric context, these are inherently preserved in the learned latent space. In contrast, SimCLR and BYOL focus on learning invariant features through contrastive objectives, without explicit constraints to retain spatial or structural information, making them less effective in encoding the clinically relevant features needed for robust classification in complex medical imaging tasks.

\subsection{Analyzing Representations}
Table \ref{tab:results4} shows classification results using an SVC on principal components of learned representations, maintaining 0.85 and 0.90 variance ($p$). This means that for evaluation on SVC, instead of inferring the classification with complete latent vector, only the elements of the latent vector which help retain $p$ amount of its variance are used. With this method, a significant deviation from the results obtained by using entire representations (Table \ref{tab:results3}) can be noticed. The difference in performance indicates that the principal components fail to preserve the structural information embedded in the representation vector, and the representation vector is not linearly separable. This result can be confirmed with a general observation on self-supervised methods. Ruff \emph{et al.} \cite{repranm} suggest that such an issue arises in self-supervised methods, since, unlike discriminative methods, self-supervised methods do not have any incentive to localize task-relevant information in selective neurons against distributing it. Combining it with results on R2RG, this suggests that the representations learned by R2RG are distributed in the latent vector, more than in the comparative methods.
Hence, a higher difference is observed. Therefore, it can be inferred that its further decomposition leads to a loss of meaningful information. Moreover, our experiments with UMAP, t-SNE, and PCA reported in Fig. \ref{fig:comps} show that this observation is consistent across different methods of extracting projections; thus, the representations are entangled.

\begin{table}[h]
    \centering
    \renewcommand{\arraystretch}{1.15}
    \caption{Ablation results of Mean Structural Similarity Index Measure (MSSIM) and PSNR scores on AOMIC, ADNI1 and PPMI datasets, with removal of some components of R2RG.}
    \begin{tabular}{p{2.cm}>{\centering\arraybackslash}p{.66cm}>{\centering\arraybackslash}p{.66cm}>{\centering\arraybackslash}p{.66cm}>{\centering\arraybackslash}p{.66cm}>{\centering\arraybackslash}p{.66cm}>{\centering\arraybackslash}p{.66cm}}
    \hline
         Method & \multicolumn{2}{c}{AOMIC} & \multicolumn{2}{c}{ADNI1} & \multicolumn{2}{c}{PPMI} \\
          & MSSIM & PSNR & MSSIM & PSNR & MSSIM & PSNR \\
         \hline
         R2RG (Vanilla Discriminator) & 0.888 & 25.46 & 0.853 & 21.42 & 0.780 & 21.24 \\
         R2RG (only $L_2$) & 0.923 & 29.06 & 0.899 & 24.16 & 0.798 & 22.81 \\
         R2RG (only $L_1$) & 0.918 & 27.51 & 0.896 & 23.72 & 0.827 & 23.11 \\
         R2RG & \textbf{0.947} & \textbf{29.11} & \textbf{0.921} & \textbf{27.14} & \textbf{0.834} & \textbf{23.24} \\
         \hline
    \end{tabular}
    \label{tab:results2}
\end{table}

\subsection{Ablation Study}
\label{sec62}
We ablate the loss function and discriminator architecture in Table \ref{tab:results2} to highlight their roles. Notably, the introduction of the PatchGAN discriminator produces substantial improvements with an improvement in both MSSIM and PSNR scores of more than 5\% across all datasets, particularly demonstrating enhanced generalization on the diverse PPMI dataset. 

Our experiments show that the combination of $L_1$ and $L_2$ is crucial for leveraging the complementary advantages of the PatchGAN discriminator, excluding either loss results in a noticeable performance drop. Additionally, a qualitative assessment of reconstruction results, illustrated in Figure \ref{fig:out_imgs}, supports the quantitative findings, indicating that the composite loss formulation achieves higher visual fidelity compared to training R2RG without $L2$ loss. Altogether, these results validate the efficacy and robustness of our design choices for the reconstruction task. Although we did not optimize for the values of $\lambda_1, \lambda_2$, we found that in a local neighborhood search, for composite loss, the values $\lambda_1 = \lambda_2 = 5$ produce better results.

\subsection{Computational Analysis}
\label{sec65}
Performing operations on 3D data is computationally expensive. However, R2RG maintains a balance between computational efficiency and model performance. As shown in Table \ref{tab:results5}, the inference time required to generate a representation vector is considerably lower than that of the baseline approaches. Moreover, despite the complexity of the architecture, R2RG inference remains comparatively efficient with 43 million parameters compared to contrastive learning baselines. This efficiency does not compromise the model’s performance, making it scalable and suitable for a wide range of applications.

\begin{table}[h]
    \centering
    \renewcommand{\arraystretch}{1.}
    \caption{Comparison of inference time and number of parameters for computational efficiency on a single example representation generation.}
    \begin{tabular}{m{1.7cm}>{\centering\arraybackslash}p{2.2cm}>{\centering\arraybackslash}p{3.5cm}}
    \hline
         Method & Inference time (s) & Number of parameters \\
         \hline
         SimCLR & 1.16 & {66,057,128} \\
         BYOL & 1.43 & \textbf{33,365,200} \\
         Ours & \textbf{0.92} & {43,785,896} \\
         \hline
    \end{tabular}
    \label{tab:results5}
\end{table}

\subsection{Limitations and Future Work}
\label{sec64}
Our analysis on latent representations showed that latent space is highly entangled and is difficult to interpret, despite superior performance in downstream classification. Thus, enhancing the interpretability of the learned representations could facilitate the real-world deployment and help end-users better understand the underlying process.  

Introducing larger unlabeled datasets can improve the network's reconstruction capability, as demonstrated in case of the larger AOMIC dataset. In addition, using a diverse dataset that includes samples from various ethnicities, sexes, and age groups would further validate generalization of the method.

\section{Conclusion}
\label{sec7}

In this paper, we study representation learning for the data-scarce 3D MRI domain. 3D MR imaging is particularly challenging due to monochromatic, high-dimensional volumes, and sparse features. We approach the problem through reconstruction, presenting R2RG, a GAN-based self-supervised learning method that reconstructs 3D volumes from latent representation vectors. R2RG unifies an autoencoder as a generator and a PatchGAN-based discriminator that provides rich spatial feedback, enhancing reconstruction fidelity and spatial understanding of the input. Our experiments demonstrate that R2RG establishes a strong reconstruction benchmark, achieving MSSIM of 0.94 on AOMIC, while generalizing effectively to smaller and more diverse datasets. This enhancement in reconstruction contributes to richer representations, improving performance on downstream classification tasks with fewer labeled examples, and outperforming contrastive-based methods particularly in smaller dataset regimes. To the best of our knowledge, this is one of the first works to focus on learning deep representations of 3D MR images. We hope that it will interest researchers in further developing the field, potentially leading to direct applications in clinical medicine with minimal labeled data.

\bibliographystyle{IEEEtran}
\bibliography{bib_final}

\end{document}